\documentclass[pdflatex,sn-basic]{sn-jnl}
\usepackage{amssymb,amsmath,mathtools}
\usepackage{graphicx}
\usepackage{algorithm}
\usepackage{algorithmic}
\usepackage{multirow,booktabs,array,adjustbox}
\usepackage{url}
\usepackage{caption}
\usepackage{rotating}
\usepackage{placeins}
\usepackage{xfp}
\usepackage{varwidth}
\hypersetup{hidelinks}
\newsavebox{\fitbox}
\newcommand{\fitwidth}[2]{\sbox{\fitbox}{\begin{varwidth}{100cm}#2\end{varwidth}}\typeout{FITRATIO #1 \fpeval{round(min(1,\columnwidth/\wd\fitbox),3)}}\ifdim\wd\fitbox>\columnwidth\resizebox{\columnwidth}{!}{\usebox{\fitbox}}\else\usebox{\fitbox}\fi}
\newcommand{\fitheight}[2]{\sbox{\fitbox}{\begin{varwidth}{100cm}#2\end{varwidth}}\typeout{FITRATIO #1 \fpeval{round(min(1,\textheight/\wd\fitbox),3)}}\ifdim\wd\fitbox>\textheight\resizebox{\textheight}{!}{\usebox{\fitbox}}\else\usebox{\fitbox}\fi}
\usepackage{xcolor}
\allowdisplaybreaks[1]

\newcommand{\akd}{\textsc{AKD}}
\newcommand{\cas}{Capability Assessment Suite}
\newcommand{\kbs}{KBI}

\begin{document}

\title[Intrinsic Confidence in Retrieval-Dominated Extractive QA]{Intrinsic Sequence-Likelihood Confidence in Retrieval-Dominated Extractive
QA: Two Pre-Specified Negatives, and What They Do and Do Not Attribute}

\author[1]{\fnm{Gunwoo} \sur{Lee}}
\author*[1]{\fnm{Changmin} \sur{Sung}}\email{shade@kisti.re.kr}
\author[1]{\fnm{Sang-Hwan} \sur{Gwak}}
\author[1]{\fnm{InA} \sur{Kim}}
\author[1]{\fnm{Ji-Young} \sur{Choi}}
\author[1,2]{\fnm{Kyong-Ha} \sur{Lee}}
\affil*[1]{\orgdiv{Large-scale AI Research Center}, \orgname{Korea Institute of Science and Technology Information (KISTI)}, \orgaddress{\street{245 Daehak-ro, Yuseong-gu}, \city{Daejeon}, \postcode{34141}, \country{Republic of Korea}}}
\affil[2]{\orgdiv{Department of Applied AI}, \orgname{University of Science and Technology (UST)}, \orgaddress{\street{217 Gajeong-ro, Yuseong-gu}, \city{Daejeon}, \postcode{34113}, \country{Republic of Korea}}}

\abstract{In extractive document question answering whose questions were generated from the passages that
contain their answers --- so that retrieval recovers 92--99.8\% of what any mode combination could
reach, whatever its absolute accuracy --- confidence-driven mechanisms have little to gain. Fine-tuning
an open language model on a specialized domain corpus yields a model whose own confidence is a tempting
control signal: it could decide which queries warrant further adaptation, and which answers to trust. We evaluate both uses under criteria fixed before the runs were executed, across four 7--9B model families whose adaptation moved closed-book F1 by at most $+0.03$, and both fail: a
distillation trigger on all four families, under its pre-specified three-step transfer budget, and a
routing-and-abstention policy in its single-model pilot.
Retrieval alone recovers 92--99.8\% of best-case combined accuracy under every
correctness criterion we test, leaving routers no meaningful gain. The sequence-likelihood signal is insufficient relative to that mode --- area under
the receiver operating characteristic curve 0.65--0.81 under the registered criterion --- before
adaptation as well as after, unchanged by scalar recalibration and not consistently improved by
token-level temperature rescaling. And the finer diagnostics depend on the correctness criterion and on
answer length; on the three adapted combinations where we could test it, selector ablations show no
statistically detectable downstream benefit from the confidence term on any seed; on Gemma, removing it
changes the selector from failing to passing both registered criteria. The usable product is a set of
pre-specified negatives with their dependencies made explicit.}


{\renewcommand{\thefootnote}{}\footnotetext{Submitted to \emph{Applied Intelligence} (APIN-D-26-13306).}}
\maketitle

\section{Introduction}
\label{sec:intro}
An enterprise that fine-tunes an open-source LLM on its own corpus acquires, along
with the adapted weights, a tempting by-product: the model's confidence in its own
answers. Two deployment decisions naturally reach for it. During continued adaptation,
low confidence flags the queries the model has not yet mastered --- candidates for
further distillation. At inference, confidence decides which answering mode should answer a
query --- the adapted parameters or a retrieval system --- and when the deployment should
decline to answer at all. Both treat the same quantity --- an intrinsic sequence-likelihood score
--- as a control signal, and both are attractive
precisely where they are hardest to check: in closed environments with
no external source of truth, using the signal is cheap and trusting it is
consequential. This paper asks whether the signal deserves that trust. This paper evaluates that operational utility: whether the score,
obtained without any additional model, can be trusted to select queries for further adaptation and to choose the
answering mode at inference.

The question is not settled by what is known about
unadapted models: the reported calibration of large models
\citep{kadavath2022language} concerns a different scale and evaluation distribution
--- we ask about 7--9B open models on a domain corpus, before and after
fine-tuning on it (Section~\ref{sec:related}). An extensive selective-prediction
literature builds routing and abstention on the premise that confidence ranks
correctness; whether that premise holds for this model class and setting has not, to
our knowledge, been tested under criteria fixed in advance.

We therefore evaluate both uses of deployed-model confidence under criteria fixed
before the corresponding runs were executed. The testbed is enterprise-style, source-grounded extractive document QA over two
corpora (with a third as a contrast set), spanning four open-source model families in
the 7--9B class: LLaMA-3.1, Qwen-2.5, Gemma-2, and Mistral. The first use is a
distillation trigger that selects, by a confidence-bearing composite score, the
queries on which further adaptation should concentrate; the second is a
routing-and-abstention policy that decides, per query, whether the adapted parameters
or retrieval should answer, and when to abstain. For the trigger, each family receives an independent verdict against the same
pre-specified criteria; routing is a single-model pilot. One caution frames everything that
follows: the two failures we report are not two pieces of evidence about the
signal; Section~\ref{sec:mechanism} separates what each is about.
Pre-specification is what makes a negative answer credible, since a post-hoc negative would always be open to the
suspicion of selective reporting. Throughout, ``registered'' means committed to version control before the corresponding run, not filed with an external registry (Online Resource 3, Section~S4 and Table~S-9).

Both uses fail their pre-specified criteria --- the
trigger on all four families, the routing policy in its single-model pilot. The
confidence signal is insufficient relative to a retrieval mode at 0.73--0.85
accuracy --- weak in that sense, the sense in which the word is used below: AUROC 0.65--0.81 across combinations (with a descriptive expected calibration error of 0.48--0.72 on the probability scale), before adaptation as well as after, and three alternative
intrinsic signals (minimum token probability, mean entropy, top-1/top-2 margin) fall
in the same band. But the failures do not reduce to the signal. Under the registered criterion the
trigger's selection concentrates the queries it targets; under token-F1 it does
not on four of five combinations, and the target is nearly the same set as `retrieval
right' either way, so we do not attribute the failure to the signal or exonerate
it, except on the three adapted combinations, where
removing the confidence term yields no statistically detectable loss and, on
Gemma, turns the
registered selector from failing both criteria to passing them
(Section~\ref{sec:mech_select}).
The routing policy is worse than always-retrieval because retrieval alone
recovers 99.3--99.8\% of the best-case combined accuracy, a ceiling no router can gain against.
Scalar recalibration cannot change the rank-based verdicts (it removes the
probability-scale offset), and token-level
temperature rescaling, which does reorder queries, improves discrimination on no combination
at $T{=}0.5$ and on a minority at $T{=}2$ or $5$.

This paper makes three contributions:
\begin{enumerate}
\item \textbf{A setting characterization robust to the correctness criterion.}
   Retrieval recovers 92--99.8\% of the best-case combined accuracy (a ratio, not an accuracy)
   under containment, token-F1 at
   two thresholds, and exact match, and the parametric mode answers correctly at most 3.4\%
   of queries under any of them --- including on the retrieval-failure subset.
\item \textbf{A signal profile with its sensitivities.} Six rank- and
   probability-scale metrics, before and after adaptation, four intrinsic signals;
   discrimination is weak under the registered criterion and weaker under
   token-F1; target over-representation holds under the registered criterion and not
   under token-F1; we therefore do not claim causal attribution. Rank-invariance
   under scalar recalibration is reported as a sanity check, not a contribution.
\item \textbf{Pre-specified negatives, with attribution not claimed except where a selector ablation resolves it.} The trigger fails criteria
   fixed before observation on all four families tested; the routing pilot
   fails them on LLaMA; two combinations are the base model
   after rollback; on one exploratory LLaMA/SciTech combination, a tenfold budget makes
   it worse.
\end{enumerate}

Two boundaries govern the scope of every claim. First, the setting: extractive,
source-grounded QA in which retrieval is structurally favored; the routing ceiling is
a finding \emph{about this setting}, and the value of confidence-based routing where
retrieval is weak or unavailable remains an open question we do not answer. Second, the signal: the verdict concerns sequence-level intrinsic
confidence --- the geometric-mean token probability and three relatives ---
separately trained estimators are outside our question.
Section~2 situates the work; Section~3 defines the signals under test; Section~4
fixes the setup; Sections~5--6 state the trigger and judge it; Sections~7--8 do the
same for routing and measure the ceiling it faces; Section~9 separates what the
failures are about --- signal, mechanism, and setting --- and identifies where
attribution is and is not supported; Sections~10--12 discuss,
bound, and conclude.

\section{Related work}
\label{sec:related}

\textbf{Selective prediction, calibration, and confidence
in language models.} Selective prediction
formalizes abstention as prediction with a reject
option~\citep{geifman2017selective}, evaluated by
risk--coverage behavior~\citep{elyaniv2010foundations}; a
parallel line studies neural-network calibration
directly~\citep{guo2017calibration}, with temperature
scaling as its standard post-hoc repair. For language
models, \citet{kadavath2022language} report that models
are, in aggregate, reasonably calibrated about what they
know on standard benchmarks. Our measurements concern a
related confidence property under different conditions --- small open
models fine-tuned on a narrow domain corpus and
evaluated on it. There the signal is weak, before
adaptation as well as after
(ECE 0.48--0.72 and correlation with correctness $+0.12$ to
$+0.32$ across four families and two corpora;
Section~\ref{sec:mechanism}). We interpret this as bounding
where that conclusion can be relied upon, not as
contradicting it
(Section~\ref{sec:discussion_principle}). Distinctly from
post-hoc calibration studies, our verdict is rank-level:
Section~\ref{sec:mech_recal} shows that no scalar monotone
recalibration can alter it and that token-level rescaling
does not consistently improve discrimination.

\textbf{Estimating LLM uncertainty beyond sequence
likelihood.} A separate line estimates uncertainty from
generation behavior rather than a single sequence score:
semantic uncertainty clusters meaning-equivalent samples
before measuring entropy~\citep{kuhn2023semantic}, with
semantic entropy applied to hallucination
detection~\citep{farquhar2024detecting};
models can be taught to verbalize
uncertainty~\citep{lin2022teaching}, prompted for
calibrated confidence~\citep{tian2023just}, and evaluated
on whether elicited confidence is
faithful~\citep{xiong2024can}; sampling-based consistency
serves as a reference-free reliability
check~\citep{manakul2023selfcheckgpt}, building on
self-consistency decoding~\citep{wang2023selfconsistency};
selective classification provides the classical
framing~\citep{geifman2017selective}. These estimators are
outside our pre-specified scope, which concerns the
intrinsic sequence-level signal a deployment uses for
free; whether they behave differently under domain
adaptation is untested here and worth testing.

\textbf{Distillation between retrieval and parametric
modes.} Knowledge
distillation~\citep{hinton2015distillation} transfers
behavior from a teacher to a student, extended to dataset
distillation~\citep{wang2018datasetdistill} and
self-distillation~\citep{zhang2019self}; a separate line integrates retrieval
with training rather than
inference~\citep{borgeaud2022retro,shi2024replug,izacard2022few}.
We are not aware of prior work that closes this loop
between two \emph{already-adapted} modes in a
post-adaptation enterprise setting, with selection driven
by the adapted model's own confidence. The trigger of
Section~\ref{sec:trigger_method} instantiates that design,
and we report it as a hypothesis that did not pass its
own pre-specified test. Self-distillation and its
relatives assume a model's own outputs carry a usable
training signal; Sections~\ref{sec:exp_trigger}
and~\ref{sec:mechanism} measure what happens when that
assumption meets an adapted model in this setting.

\textbf{Routing between retrieval and parametric answering.} Adaptive retrieval decides during generation when to
consult a retriever: FLARE retrieves again whenever a tentative next sentence contains low-confidence
tokens~\citep{jiang2023flare}, and Adaptive-RAG trains a classifier of question complexity that chooses among no
retrieval, single-step retrieval, and multi-step retrieval~\citep{jeong2024adaptiverag}. Whether to retrieve at all has
been learned through reflection tokens~\citep{asai2023selfrag} and tied to the popularity of the queried entity, with
retrieval helping most on less popular knowledge~\citep{mallen2023trust}. Among whole models, RouteLLM trains routers on
preference data to send each query to a stronger or a weaker language model~\citep{ong2024routellm}. Applied work
strengthens the reliability of question answering by extending retrieval with knowledge graphs~\citep{linders2025kgrag}
and by examining whether systems learn to recognize unanswerable questions~\citep{reyesmontesinos2025answerability}.
These works assume there is a gain to route into; Section~\ref{sec:ablation} measures it first.

\section{Signals under test}
\label{sec:kbd_signals}\label{sec:signals}

Three quantities recur in everything that follows. Two are
per-query signals consumed by both the trigger and the
router; the third is a corpus-level diagnostic the router
uses as a prior. Their reliability turns out to determine what the components built on
them can achieve.

\textbf{Parametric confidence $p_{\mathrm{FT}}(q)$} is the
geometric mean of the generated tokens' probabilities ---
$\exp$ of the mean token log-probability over the generated
sequence, prompt tokens excluded, the end-of-sequence token
included in the likelihood though not in the returned text.
It is length-normalized by construction.
$p_{\mathrm{FT}}$ is a length-normalized sequence
likelihood, not an estimate of answer probability; we
therefore treat probability-scale metrics (ECE, Brier, NLL)
as descriptive summaries of its scale, and rest comparative
claims on rank-based metrics (AUROC, AURC, selective
accuracy). This is the signal
named in the title; Section~\ref{sec:mechanism} evaluates
it alongside three other intrinsic signals computed from
the same decoding pass. Its reliability is poor on the recorded held-out query sets
--- the SciTech half predating the template fix of Section~\ref{sec:setup}: over the
LLaMA deployment's
$n = 2{,}860$ combined SciTech and WikiGen held-out
queries the signal is both badly shifted and
weak: mean confidence exceeds accuracy by $+0.575$, and its
correlation with correctness is $+0.208$ (95\% CI
$[+0.18, +0.24]$). The reliability diagram is weakly
monotone and of little use against a retrieval mode at 0.73--0.85 accuracy
--- accuracy rises
from 1.0\% in the lowest confidence decile to only 24.1\%
in the highest. The SciTech half of these figures is superseded for diagnostic
purposes by the regenerated set of Online Resource 3, Table~S-18. These pooled two-corpus figures are a
different statistic on a different set than the per-family
binned calibration errors of Section~\ref{sec:mechanism},
and the two should not be interpreted as the same number.

\textbf{Retrieval context similarity $\phi(q)$} is the cosine
similarity, under the deployment's single embedding model,
between the retrieval mode's answer and its retrieved
context:
$\phi(q) = \cos(\mathbf{e}_{A_{\mathrm{RAG}}(q)},
\mathbf{e}_{c_{\mathrm{RAG}}(q)})$. It is an embedding
proxy for whether the answer is grounded in the retrieved context --- high when the answer restates
content present in the retrieved passage --- and
deliberately not an LLM-judge score: a local judge would
add a separate judge-based reliability signal, outside the present scope. The proxy is correspondingly
coarse and we treat it as such
(Section~\ref{sec:discussion_limitations}).

\textbf{The knowledge-coverage diagnostic (KBI)} is a
corpus-level scalar summarizing how much of a corpus the
base model \emph{already covers}: the mean similarity, over
$K = 50$ closed-book concept probes generated from the
corpus before adaptation, between the base model's answer
and the source passage --- higher KBI means greater overlap
with parametric knowledge (construction in
Online Resource 3, Section~S1; the companion study\footnote{A companion manuscript develops the governance components referenced here.} develops it as an adaptation-risk diagnostic). Its only
role in this paper is the routing prior of
Section~\ref{sec:routing}.

Both per-query signals are consumed as \emph{relative}
orderings, through percentile ranks; neither supports an
absolute decision threshold.

\section{Experimental Setup}
\label{sec:setup}

\textbf{Base models.} All experiments run on four
instruction-tuned open models from four architecture
families: LLaMA-3.1-8B-Instruct, Qwen2.5-7B-Instruct,
Gemma-2-9b-it, and Mistral-7B-Instruct-v0.3. No model is
accessed as a hosted service. Coverage is deliberately
uneven and we state its shape at the outset: the trigger verdict
runs on all four families over SciTech, with WikiGen added
for LLaMA; the routing-and-abstention pilot is reported for LLaMA on both
corpora; the trigger's mechanism diagnostics cover all four families
on SciTech from the same recorded artifacts, and LLaMA alone on
WikiGen (Section~\ref{sec:mechanism}); the
mode ablation that establishes the routing ceiling
covers all four families. Each result names the models and
corpora it rests on.

Two of the five evaluated combinations are not, in fact, adapted.
On Mistral/SciTech and LLaMA/WikiGen the deployment's
capability guard rolled adaptation back at its first
evaluation, leaving an identity adapter; every ``adapted''
measurement on those combinations --- trigger verdict,
diagnostics, and the full-fine-tuning comparison's LoRA
reference --- therefore concerns the base model. We label
them as such throughout and count three genuinely adapted
combinations (LLaMA, Qwen, Gemma on SciTech).

\textbf{Corpora.} Two domain corpora provide the evaluation:
SciTech (500 English scientific and technical documents)
and WikiGen (1{,}000 English general-encyclopedic
documents). NaturalQuestions contributes 37 held-out examples drawn from its validation split as a high-overlap contrast set
(the 15\% validation partition of a 250-example sample under the corpus-stratified 70/15/15 document split), used only
in the pre-/post-adaptation comparison of
Section~\ref{sec:mechanism}. The corpora are public
collections --- arXiv-derived technical documents, statute
text, and encyclopedia articles --- serving as surrogates
for enterprise document sets; the deployment constraints,
not the documents themselves, are what is proprietary about
the setting.

\textbf{Preparation and splits.} Documents are chunked at
512 tokens with overlap 128. QA pairs are generated from
each chunk with the hosted model of the Correctness paragraph below at dataset-construction
time and partitioned at
\emph{document} level into training, retention, and test
splits under an automated assertion that no document
contributes to more than one split; measured
query-set--test overlap is zero. The held-out query sets hold
1{,}196 queries (SciTech) and 1{,}759 (WikiGen), identical
across base models; the mode ablation evaluates their
combined 2{,}860 gold-span-eligible queries, and the
retrieval-displacement analysis uses hard-negative subsets
of 386 and 318.

\textbf{Adaptation and evaluation.} Fine-tuning is LoRA
(rank 16, scaling 32, dropout 0.05) on all seven attention
and MLP projections over a 1{,}000-step budget. Domain
performance is token-level F1 on held-out QA, closed-book
or with retrieval as indicated. The trigger experiment
repeats over seeds 42, 43, and 44 and reports every seed
rather than their mean. A capability-regression guard
(Section~\ref{sec:trigger_method}) monitors a fixed
general-capability benchmark set during adaptation; its
contribution is not a claim of this paper. Hardware and the
pinned software stack are in Online Resource 3.

\textbf{Correctness and eligibility.} An answer is correct
if it contains the recorded gold span; where a query has no
recorded short answer, a token-F1 of at least 0.3 against
the reference substitutes. Of the 2{,}955 combined
held-out queries, 2{,}860 are gold-span-eligible (a
recorded short answer exists); the 95 remainder are
excluded from span-based statistics. QA pairs were
generated at dataset-construction time by OpenAI's gpt-5.5 (the served alias;
no dated snapshot identifier was recorded), as structured-output JSON at
temperature 0.2 (two to four per chunk under a seven-rule extractive-span
prompt, 3.8 on average) under the prompt version of the released snapshot,
unchanged since the generation run,
then filtered for question length and self-reference, with
any short answer that is not a verbatim substring of its
chunk nulled. The generation step is therefore not reproducible bit-for-bit; every generated artifact used in the
experiments is preserved and released in the data record (Online Resource 2). Section~\ref{sec:mech_criterion} reports how the
diagnostics move under token-F1 and exact-match criteria; the
trigger verdicts of Section~\ref{sec:exp_trigger} use token-F1 on held-out test
sets; the routing pilot of Section~\ref{sec:routing} was evaluated on the held-out
query sets under the registered containment criterion, and Section~\ref{sec:mech_criterion}
shows its ceiling under every criterion; Online Resource 3, Table~S-8 gives the
split, criterion, and resampling unit of every registered verdict.

\textbf{One recorded combination predates a fix.} The LLaMA/SciTech
held-out query set was built, before a template correction, with a
Llama-2-style prompt template that the Llama-3.1 tokenizer does not
terminate, so both modes' answers ran to the 64-token limit (median
45 words); the other four combinations postdate the correction (median 2--7
words). We keep the recorded combination as the registered
artifact --- the trigger verdict on it consumed these signals --- and
report regenerated closed-book diagnostics under the corrected prompt template
alongside (Online Resource 3, Table~S-6(c)); we regenerated it under the
corrected prompt template and re-ran the trigger and the routing pilot on it as a
post-registration replication (Online Resource 3, Section~S5); the registered artifact
and its verdicts stand as recorded, and the replication is reported beside them. We retain the registered artifact for
protocol transparency and use the corrected-template replication for substantive interpretation.

\textbf{Retrieval quality.} Recomputed deterministically
over the released index, the retriever places the gold
chunk at rank 1 for 48.7\% of SciTech queries and 84.7\%
within the top 4 (median rank 2); on WikiGen, 72.1\% and
96.1\% (median rank 1). On the stored hard-negative
subsets the recorded gold rank has median 2 on both corpora
(90th percentile 6 and 3). The routing ceiling of
Section~\ref{sec:ablation} sits on top of this:
retrieval's dominance is a property of the mode, not of
an unusually easy index.

\textbf{Small adaptation effect.} Adaptation moved closed-book F1 by
$+0.017$ (LLaMA), $+0.031$ (Qwen), $+0.010$ (Gemma) over the base
models (Online Resource 3, Table~S-12); on Mistral/SciTech and
LLaMA/WikiGen the guard rolled adaptation back and the deployed model
is the base model. The parametric mode therefore operates near its
floor on every combination, a property that governs the calibration and
discrimination figures below.

\textbf{One property of this setting governs every result.}
The QA is \emph{extractive and source-grounded}: the answer
to a generated question is present in the chunk it was
generated from, so retrieval that surfaces the right chunk
surfaces the answer, and retrieval-side scores are high
throughout. This is realistic for enterprise-style, source-grounded extractive document QA --- and a setting in
which retrieval holds an intrinsic advantage. The routing ceiling of
Section~\ref{sec:ablation} and the scope of every
conclusion in this paper are conditioned on it; we flag the
dependence where it binds and return to it in the
limitations.

\section{Method under test: the distillation trigger}
\label{sec:trigger_method}
\label{sec:akd}\label{sec:akd_trigger}

The first use of deployed-model confidence we evaluate is a
distillation trigger (\akd{} in the released artifacts): a
mechanism that selects, from a
held-out query set, the batch on which
further adaptation should concentrate. The premise is the
standard one: a query is a useful distillation target when
the retrieval mode and the parametric mode disagree
\emph{and} one of them is clearly the more reliable.
Confidence enters as the parametric mode's
reliability estimate, which is exactly the role our verdict
examines.

For each retained query $q$, the trigger computes three
signals: the mode disagreement
\begin{equation}
  d(q) = 1 - \cos\!\bigl(\mathbf{e}_{A_{\mathrm{RAG}}(q)},\;
                         \mathbf{e}_{A_{\mathrm{FT}}(q)}\bigr),
  \label{eq:signal_d}
\end{equation}
the retrieval context-similarity proxy $\phi(q)$, and the
parametric confidence $p_{\mathrm{FT}}(q)$ (both defined in
Section~\ref{sec:signals}); answer embeddings come from the
single encoder used throughout. Each signal is normalized to its percentile rank within the held-out
query set, and the ranked signals combine into two
directional composite scores,
$S_{\mathrm{R}\to\mathrm{F}}$ preferring a retrieval answer that
agrees with its retrieved text over an under-confident parametric answer
and $S_{\mathrm{F}\to\mathrm{R}}$ the converse; both share
the disagreement term (Online Resource 3, Section~S3 states
them in full). Selection takes the top-$N$ queries per
direction ($N = 32$ throughout), a budget-based rule that
guarantees non-degenerate batches; an earlier
threshold-conjunction gate selected nothing on held-out query sets of
this size and is not the path evaluated here
(Online Resource 3, Section~S3).

Each refinement cycle applies one LoRA step to the
retrieval-to-parametric batch under a task--distillation
objective that pulls the parametric response toward what
the same model produces with retrieved evidence in context
(Online Resource 3, Section~S3, Eq.~S6);
the reverse batch is added to a retrieval preference set
that reranks candidate chunks toward agreement with
high-confidence parametric outputs, leaving model weights
untouched. All three signals, the composite scores, and
the budget are computed once at loop entry and reused
across cycles --- a design whose consequences the mechanism
analysis in Section~\ref{sec:mechanism} examines. A
capability-regression guard follows each cycle, rolling the
adapters back if a fixed general-capability benchmark
set degrades beyond a preset margin; the guard was
active in every run but never separately ablated, and we
make no claim about its contribution. A per-cycle rate cap
tied to a corpus-level diagnostic band exists in the
implementation but was non-binding in every reported run
--- both evaluated corpora fall in the band assigning the
cap its highest value, which the fixed budget never reaches
--- so no result depends on it
(Online Resource 3, Section~S3).

Two scope statements bound what the verdict can mean. The
loop is an offline refinement procedure between two
already-adapted modes, operating on the held-out query set
rather than on live queries; and that held-out query set is
disjoint from the test set at document level
(Section~\ref{sec:setup}). The mechanism is specified precisely enough that its pre-specified
failure (Section~\ref{sec:exp_trigger}) can be localized rather than
merely observed (Section~\ref{sec:mechanism}).

\section{Experiment: the trigger verdict}
\label{sec:exp_trigger}\label{sec:exp_akd}

This experiment judges the trigger of Section~\ref{sec:trigger_method}
against criteria fixed before the runs. Table~\ref{tab:akd_criteria}
reports the outcome per family: \emph{paper scope} marks a mechanism
rejected and reported as a negative result, and \emph{conditional} marks
R-1 passed but R-2 failed, rejected by the pre-fixed rule.

\textbf{Pre-specified criteria.} Five criteria were
registered, three for the retrieval-to-parametric direction
and two for its converse. \textbf{R-1}: distilled
closed-book F1 exceeds the undistilled model on a majority
of seeds with bootstrap $P > 0.7$ on a majority of seeds.
\textbf{R-2}: the same improvement over a budget-matched
\emph{random} selection, controlling for the effect of
training on additional queries at all. \textbf{F-1}: the
converse direction produces a non-degenerate preference
set. \textbf{F-2}: injecting those preferences improves the
rank of gold evidence; \textbf{F-3}: that improvement
holds under bootstrap resampling. The decision rule was fixed
with the criteria: failure of R-2 or F-2 rejects the
mechanism regardless of the other outcomes, since these are
the controls separating a real effect from an artifact of
extra training or of arbitrary preference injection.
Scale-up used 32 queries per direction, three cycles, seeds
42/43/44, on the SciTech held-out query set for all four
families, with a WikiGen scale-up for LLaMA.

\begin{table}[t!]
\centering
\scriptsize
\caption{Pre-specified criteria across four base models.
\emph{paper scope}: mechanism rejected, reported as
negative result; \emph{conditional}: R-1 passed but R-2
failed, rejected by the pre-fixed rule. The replication row is post-registration
(Online Resource 3, Section~S5); its R-2 pass does not change the verdict, which the
pre-fixed rule ties to R-1 and R-2 jointly. Effect sizes in
Table~\ref{tab:effect}; selector ablations that pass the same criteria without
the confidence term are in Online Resource 3, Table~S-21.}
\label{tab:akd_criteria}
\setlength{\tabcolsep}{4pt}
\fitwidth{tab:akd_criteria}{%
\begin{tabular}{lccccccl}
\toprule
\textbf{Base model} & \textbf{R-1} & \textbf{R-2}
  & \textbf{F-1} & \textbf{F-2} & \textbf{F-3}
  & \textbf{Pass} & \textbf{Verdict} \\
\midrule
LLaMA-3.1-8B & $\times$ & $\times$ & \checkmark & $\times$ & $\times$
  & 1/5 & paper scope \\
Qwen2.5-7B   & \checkmark & $\times$ & \checkmark & $\times$ & $\times$
  & 2/5 & conditional \\
Gemma-2-9B   & $\times$ & $\times$ & \checkmark & $\times$ & $\times$
  & 1/5 & paper scope \\
Mistral-7B   & $\times$ & $\times$ & \checkmark & $\times$ & $\times$
  & 1/5 & paper scope \\
\addlinespace
LLaMA-3.1-8B (corrected-template replication) & $\times$ & \checkmark & --- & --- & ---
  & --- & paper scope (R-1 fails) \\
\bottomrule
\end{tabular}%
}
\end{table}

\begin{sidewaystable}
\centering
\scriptsize
\caption{The trigger verdicts as effect sizes: paired closed-book test-F1
differences of the distilled model against the undistilled model (R-1) and
against a budget-matched random batch (R-2), per family and seed, with 95\%
bootstrap intervals by query and by document cluster ($B = 10{,}000$). The
registered-form intervals ($N = 1{,}000$) are in Online Resource 3, Table~S-8.}
\label{tab:effect}
\setlength{\tabcolsep}{5pt}
\fitheight{tab:effect}{%
\begin{tabular}{llcccc}
\toprule
 & & \multicolumn{2}{c}{\textbf{$\Delta$F1 vs.\ Off (R-1)}} & \multicolumn{2}{c}{\textbf{$\Delta$F1 vs.\ Random-$N$ (R-2)}} \\
\textbf{Model / corpus} & \textbf{Seed} & \textbf{query} & \textbf{cluster} & \textbf{query} & \textbf{cluster} \\
\midrule
LLaMA-3.1-8B / SciTech & 42 & $-0.0045$ [-0.0131, +0.0040] & $-0.0045$ [-0.0135, +0.0044] & $-0.0064$ [-0.0117, -0.0013] & $-0.0064$ [-0.0118, -0.0012] \\
 & 43 & $-0.0044$ [-0.0129, +0.0041] & $-0.0044$ [-0.0139, +0.0050] & $-0.0079$ [-0.0142, -0.0020] & $-0.0079$ [-0.0145, -0.0014] \\
 & 44 & $-0.0046$ [-0.0131, +0.0040] & $-0.0046$ [-0.0142, +0.0046] & $-0.0065$ [-0.0136, +0.0004] & $-0.0065$ [-0.0140, +0.0008] \\
\addlinespace
Qwen2.5-7B / SciTech & 42 & $+0.0026$ [-0.0023, +0.0075] & $+0.0026$ [-0.0019, +0.0070] & $-0.0047$ [-0.0101, +0.0006] & $-0.0047$ [-0.0096, +0.0000] \\
 & 43 & $+0.0043$ [-0.0007, +0.0094] & $+0.0043$ [+0.0001, +0.0086] & $+0.0014$ [-0.0052, +0.0081] & $+0.0014$ [-0.0048, +0.0075] \\
 & 44 & $+0.0028$ [-0.0021, +0.0078] & $+0.0028$ [-0.0012, +0.0071] & $-0.0033$ [-0.0092, +0.0026] & $-0.0033$ [-0.0092, +0.0025] \\
\addlinespace
Gemma-2-9B / SciTech & 42 & $-0.0025$ [-0.0092, +0.0040] & $-0.0025$ [-0.0103, +0.0051] & $-0.0021$ [-0.0096, +0.0052] & $-0.0021$ [-0.0100, +0.0055] \\
 & 43 & $-0.0016$ [-0.0079, +0.0047] & $-0.0016$ [-0.0085, +0.0053] & $+0.0010$ [-0.0061, +0.0083] & $+0.0010$ [-0.0068, +0.0086] \\
 & 44 & $-0.0013$ [-0.0077, +0.0049] & $-0.0013$ [-0.0088, +0.0060] & $+0.0009$ [-0.0063, +0.0082] & $+0.0009$ [-0.0073, +0.0090] \\
\addlinespace
Mistral-7B / SciTech & 42 & $+0.0011$ [-0.0063, +0.0085] & $+0.0011$ [-0.0079, +0.0098] & $+0.0012$ [-0.0054, +0.0077] & $+0.0012$ [-0.0062, +0.0084] \\
 & 43 & $+0.0022$ [-0.0051, +0.0095] & $+0.0022$ [-0.0069, +0.0109] & $+0.0054$ [-0.0016, +0.0126] & $+0.0054$ [-0.0033, +0.0145] \\
 & 44 & $+0.0017$ [-0.0058, +0.0091] & $+0.0017$ [-0.0075, +0.0107] & $+0.0010$ [-0.0062, +0.0082] & $+0.0010$ [-0.0068, +0.0090] \\
\addlinespace
LLaMA-3.1-8B / WikiGen & 42 & $-0.0007$ [-0.0054, +0.0039] & $-0.0007$ [-0.0050, +0.0036] & $-0.0022$ [-0.0063, +0.0020] & $-0.0022$ [-0.0060, +0.0016] \\
 & 43 & $-0.0008$ [-0.0058, +0.0039] & $-0.0008$ [-0.0052, +0.0036] & $-0.0018$ [-0.0069, +0.0032] & $-0.0018$ [-0.0066, +0.0030] \\
 & 44 & $+0.0002$ [-0.0047, +0.0051] & $+0.0002$ [-0.0040, +0.0045] & $+0.0023$ [-0.0026, +0.0073] & $+0.0023$ [-0.0026, +0.0071] \\
\bottomrule
\end{tabular}
}
\end{sidewaystable}

Table~\ref{tab:effect} states the same evidence as effect sizes with intervals;
the pass/fail vector is the pre-specification's record, and the reader who
prefers estimates to verdicts should read the table. The $P > 0.7$ criterion
is a registered convention, not a standard threshold, and we do not defend it
beyond its having been fixed in advance.

\textbf{Outcome.} Table~\ref{tab:akd_criteria} reports the
verdict. Under the corrected prompt template the LLaMA combination's selection exceeds the
random control but not the undistilled baseline, so its verdict is unchanged
(Table~\ref{tab:akd_criteria}, last row). Three families produce byte-identical criteria
vectors (1/5 passed; F-1 only) and Qwen differs on R-1
alone (2/5) --- the automatic verdict generator records
this as partial family invariance: from the four-family
perspective, three exact reproductions and one divergence,
and we use both formulations rather than the more
flattering one. The LLaMA/SciTech selection consumed the pre-fix retention
signals (Section~\ref{sec:setup}); the verdict is reported as recorded. The divergence does not change the
conclusion, by the decision rule fixed in advance: Qwen
fails R-2, and a method that improves on doing nothing but
not on selecting queries at random has not shown that its
selection carries information. F-2 and F-3 fail on all four
families, so the converse direction is rejected everywhere.
For Mistral, and for the WikiGen scale-up of LLaMA, the
verdict concerns the base model (Section~\ref{sec:setup}).

\textbf{Replication under the corrected prompt template.} Replicated under the corrected
prompt template, the LLaMA/SciTech trigger fails R-1 and passes R-2: the distilled model
exceeds Off on all three seeds but reaches $P > 0.7$ on none ($P$ 0.66 / 0.69 /
0.59 at query level, 0.64 / 0.66 / 0.56 by document cluster), and exceeds
Random-$N$ on all three with $P$ 0.92 / 0.84 / 0.61 (0.90 / 0.82 / 0.57 by
cluster) --- R-2 passes where the registered artifact failed ($P$ 0.01 / 0.00 /
0.03), R-1 is unchanged, and the registered verdict stands as recorded
(Online Resource 3, Section~S5). For a reader weighing evidence rather than protocol
fidelity, the corrected combination is the informative one; the registered row is
retained as the record of what was pre-specified and observed, not as the better
estimate.

\textbf{Seed-level detail.} Criteria are booleans over
three seeds, and booleans hide how close a decision was.
Online Resource 3, Table~S-2 reports every seed;
Online Resource 3, Figure~S-1 plots per-seed bootstrap probabilities
against the 0.7 line. Qwen's R-1 pass is not marginal ---
three of three seeds pass the criterion, two comfortably.
Mistral's R-1 failure is: its distilled model exceeds
baseline on all three seeds and one reaches $P = 0.726$, so
it joins LLaMA's verdict class by a very narrow margin while its
seed pattern resembles Qwen's. The three ``exact
reproductions'' are exact at the level of the criteria
vector, not of the underlying statistics. Neither
observation makes the mechanism succeed: across all twelve
model--seed pairs, no model passes R-2.

\section{The routing policy and its verdict}
\label{sec:routing}

The second use of deployed-model confidence acts at
inference time: given the fine-tuned model and a retrieval
pipeline, decide per query which mode answers, or
whether to decline. Declining carries weight where there is no escalation path --- a
confidently wrong answer is more costly than an unanswered one --- so
the policy under test is three-way: retrieval, parametric, or abstain. We report it as a pre-specified pilot that did
not pass its own acceptance criteria: two of four criteria
passed, so the component is deferred rather than claimed,
and we present it in full because its failure mode is
informative about the setting and the criteria predate the
runs.

\textbf{Policy.} Both mode scores are percentile ranks
of the per-query signals of Section~\ref{sec:signals}: the
retrieval score is the context-similarity rank
$\phi_{\mathrm{rank}}(q)$; the parametric score is the
confidence rank $p_{\mathrm{FT,rank}}(q)$ less a
corpus-level penalty $\gamma\,\mathrm{KBI}$, where KBI is
the knowledge-coverage diagnostic of
Section~\ref{sec:signals}. One inconsistency in the
registered design is reported here rather than repaired:
KBI is highest where base-model overlap with the corpus is
greatest, so the penalty as registered and executed demands
the most query-level confidence on the corpora the model
knows best, while the registration's own comment motivates
it in the opposite, coverage-gap interpretation. We evaluate the
policy exactly as registered; the E-3 verdict below
concerns the penalty as executed, and on both evaluated
corpora the corpus-level score in any case falls in the
low-KBI region (see the degeneracy note below). The query
goes to the higher score. The policy
declines in two mutually exclusive ways: on \emph{low
confidence}, when the higher score falls below
$\theta_{\mathrm{abstain}}$, and on \emph{conflict}, when
both raw ranks exceed $\theta_{\mathrm{high}}$ yet the
adjusted scores lie within $\theta_{\mathrm{conflict}}$ of
each other; the gate uses raw ranks and the margin uses
adjusted scores, since gating on adjusted scores would
leave the conflict branch unreachable (full statement in
Online Resource 3, Section~S2). In the data the conflict
branch activates rarely --- ten abstentions on SciTech, none on
WikiGen --- so low-confidence abstention accounts for
essentially all declined queries. All four constants
($\gamma = 2.0$, $\theta_{\mathrm{abstain}} = 0.3$,
$\theta_{\mathrm{high}} = 0.7$,
$\theta_{\mathrm{conflict}} = 0.15$) were fixed in the
pre-specification document and not adjusted afterwards.

\textbf{Criteria and verdict.} Four criteria were
registered, with the rule that the component is claimed
only if all four pass. \textbf{E-1} (abstention quality):
accuracy on answered queries exceeds accuracy on abstained
ones, bootstrap $P > 0.7$ (the registered threshold; the
observed probability, 1.000 on both corpora, exceeds any
conventional bar). \textbf{E-2} (risk--coverage):
the policy's risk--coverage AUC improves on both
registered baselines --- always-retrieval and a simple maximum-confidence
router --- each taken separately.
For the single-mode baselines, the risk--coverage curve orders queries by the
mode's own score --- the context-similarity rank for always-retrieval, the
$p_{\mathrm{FT}}$ rank for always-parametric, and the larger of the two for the
maximum-confidence router --- and sweeps retained coverage; their reported
selective accuracy is the full-coverage operating point.
\textbf{E-3} (prior
informativeness): routing accuracy at $\gamma = 2$ exceeds
$\gamma = 0$, bootstrap $P > 0.95$. \textbf{E-4} (ensemble
advantage): the policy outperforms each of its own
components. The verdict is two of four: E-1 passes on both
corpora with bootstrap probability 1.000, and E-3 passes by
the registered rule ($P > 0.95$ on both corpora), a pass we do not interpret as
evidence about \kbs{} (Section~\ref{sec:mech_route}); E-2 and E-4 fail.
One correction is on
the record: an initial evaluation returned three of four,
because the risk--coverage area had been computed against a
single pooled reference where E-2 as registered requires
each baseline separately; after correcting the evaluation
code E-2 fails, and we report the corrected figures rather
than replacing without note the earlier number. The pilot ran
on LLaMA-3.1-8B over both held-out query sets; it was not
extended to the other families and one registered ablation
(a retrieval-cosine variant of the RAG score) was not run
--- both because the pre-specification deferred further
investment once the bar was missed.

On the regenerated held-out query set the pilot's criteria give E-1 pass, E-2 fail,
E-3 pass, E-4 fail against the registered pass, fail, pass, fail; the vector is
unchanged (two of four), with retrieval accuracy 0.750 against 0.806, the
policy's selective accuracy 0.648 at coverage 0.80 against 0.725 at 0.81, and
E-3's $P$ at 1.000 on both corpora as registered (Online Resource 3, Section~S5).

\textbf{Why the failures are structural.} The policy must
exceed the better of its components, and
Section~\ref{sec:ablation} measures the possible gain from that at
0.002--0.005 of accuracy: best-case routing --- knowledge, after the fact,
of which mode is correct per query --- exceeds
always-retrieval by under 0.3 percentage points on either
corpus, because only 5--14 of 2{,}860 queries are answered
parametrically and not by retrieval. A realisable policy
gives up far more than that by
routing queries to the weaker mode --- 0.684 against
always-retrieval's 0.806 on SciTech, 0.746 against 0.814 on
WikiGen. The ceiling explains why the policy cannot gain, not the size of
what it gives up. E-2 and E-4 fail in a setting that
holds no possible gain for any policy of this form; whether a
better-built policy could succeed elsewhere is not tested
here. A degeneracy compounds it: both corpora
sit below the prior's lower band, so KBI-only and
always-retrieval coincide here.

\textbf{What the passes are worth.} E-3 establishes that
the registered offset improves accuracy over $\gamma = 0$;
it does not establish that KBI is informative --- both
corpora share the low-KBI region and the offset's polarity
runs opposite to its registered rationale, so the prior's
informativeness is unidentified here.
Reversing the offset's sign, as its registered rationale would
have it, lowers routing accuracy to 0.25 and 0.19 (exploratory, not a
registered criterion); the executed sign is the empirically right one.
E-1 is real but narrower than it looks: the policy is
correct on 72.5\% of answered queries against 51.1\% of
declined ones on SciTech, so abstention does order queries
by difficulty --- yet answering 80.9\% of queries at 72.5\%
accuracy is worse than answering all of them by retrieval
at 80.6\%. Online Resource 3, Figure~S-2 traces the swept
operating points; the registered operating point lies below
always-retrieval's full-coverage accuracy on both corpora,
and on WikiGen only the lowest-coverage sweep point exceeds
it --- at roughly half coverage (per-point figures in
Online Resource 3, Table~S-15). What remains after that
comparison is a rank-score calibration effect: the
abstaining policy's expected calibration error, computed on
percentile-rank scores rather than probabilities, is 0.136
against always-retrieval's 0.306 on SciTech and 0.245
against 0.332 on WikiGen, and the gap persists at matched
coverage (0.093 against 0.283; 0.164 against 0.236).
Section~\ref{sec:discussion} states what that is and is
not worth.

Percentile ranks in the pilot were computed within the
evaluation set, a transductive convenience. Fixing the
empirical CDF on a held-out half and mapping the other half
through it leaves the policy's accuracy within 0.003
(0.742 vs.\ 0.739 on SciTech, 0.785 vs.\ 0.785 on WikiGen)
at slightly lower coverage; the verdict does not depend on
the transductive step.

\textbf{Routing where there is possible gain.} Restricted to the queries whose gold
chunk is not retrieved at rank 1 (614 / 490 on SciTech / WikiGen) and beyond rank
4 (183 / 69) --- the hard-negative subsets are test-split queries and do not
intersect the held-out query sets --- the best-case-routing possible gain over always-retrieval
is 0.000--0.005; the policy's selective accuracy at its registered operating point
is 0.684 and 0.268 on SciTech (coverage 0.76 and 0.53) and 0.712 and 0.184 on
WikiGen (0.77 and 0.71) against always-retrieval's 0.707, 0.224, 0.745, and 0.217,
and its risk--coverage area 0.211 and 0.679 against 0.120 and 0.586 on SciTech,
0.242 and 0.795 against 0.219 and 0.769 on WikiGen (Online Resource 3, Section~S6,
Table~S-19). The policy exceeds always-retrieval on none of them at full
coverage (forced-routing accuracy 0.609, 0.191, 0.667, and 0.174), and with
abstention only on the SciTech beyond-rank-4 subset, by answering 53\% of its 183
queries (0.268 against 0.224): where retrieval fails, the parametric mode fails
too (accuracy 0.041--0.084), and the signal does not tell the two apart. Either
way the setting-imposed ceiling of Section~\ref{sec:ablation} is not an artifact of the
easy queries.

\section{The routing ceiling}
\label{sec:ablation}

\begin{figure}[t]
  \centering
  \includegraphics[width=\linewidth]{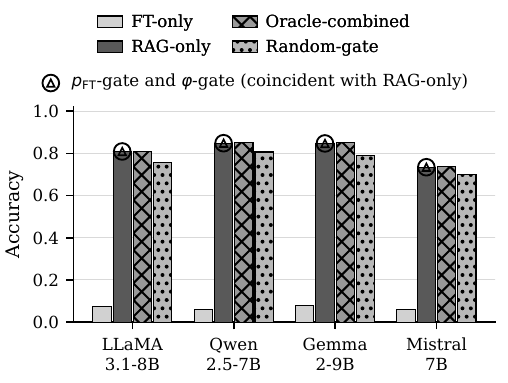}
  \caption{Component ablation across four base models.}
  \label{fig:ablation}
\end{figure}

Before judging any per-query policy, we measure what
per-query routing could gain at best. This experiment
computes the ceiling on all four base models, over the
combined SciTech and WikiGen held-out query sets
($n = 2{,}860$ per model): Figure~\ref{fig:ablation} plots
the result and Online Resource 3, Table~S-1
reports it numerically. Four findings hold across all four
model families.

\textbf{Retrieval recovers essentially the whole ceiling.}
RAG-only reaches 99.3--99.8\% of the best-case combined accuracy
(retrieval recall@4 of 0.847 and 0.961;
Section~\ref{sec:setup}),
leaving 0.2--0.7\% of accuracy for any router to compete
for. The ceiling is not an artifact of easy retrieval: restricted to
queries whose gold chunk is not retrieved at rank 1, the parametric mode is
right where retrieval is wrong on at most six queries per combination
(Section~\ref{sec:mech_criterion}). Recovery of the best-case combination is not accuracy: under
exact match retrieval answers 0.207--0.484 of queries and still recovers
94.6--99.5\% of the best-case combination, because the parametric mode adds almost nothing
under any criterion.

\textbf{The parametric mode adds almost nothing.}
FT-only accuracy is 0.058--0.081, and the fine-tuned model
is right where retrieval is wrong on only 5 to 14 of
2{,}860 queries---the same gap counted from the parametric side.

\textbf{Best-case gating saturates at always-retrieval.}
Both gates return accuracy identical to RAG-only, to
floating-point precision, on every model: the
threshold chosen with hindsight reduces to a policy that
never selects the parametric mode. The threshold chosen with hindsight is an upper bound within this single-score
threshold-gating class, not a tuning problem; the best-case combined row bounds
every policy.

\textbf{Routing at random is strictly worse.}
Random gating falls below RAG-only on every model, by 3.6
to 5.9 points (from unrounded values), in proportion to how
often the parametric
mode is mixed in.

The per-query analysis (LLaMA, on the recorded held-out query sets;
Section~\ref{sec:setup} and Online Resource 3, Section~S5) explains why: the
fine-tuned model's mean $p_{\mathrm{FT}}$ is 0.648 against an
accuracy of 0.073, a
raw score--accuracy gap of $+0.575$, and even within its most
confident 3\% of queries it reaches 42.2\% where retrieval
attains 81\%.

There is essentially no possible gain for any router. This ceiling
predicts the routing verdict of Section~\ref{sec:routing}
on its own; the trigger's verdict, as
Section~\ref{sec:mechanism} shows, has a different
explanation.

\section{Where the failures lie}
\label{sec:mechanism}

Both mechanisms having failed their criteria
(Sections~\ref{sec:exp_trigger}--\ref{sec:routing}), this section
separates three things the failures could be about --- the signal, the mechanism,
and the setting --- and shows that attribution differs across the two mechanisms and, for the
trigger, across combinations.

\subsection{The signal, profiled}
\label{sec:mech_diag}

\begin{table}[t!]
\centering
\scriptsize
\caption{Trigger diagnostics. SciTech
diagnostics are recorded per family; WikiGen diagnostics
exist for LLaMA only. The converse-direction
(parametric-to-retrieval) displacement panel is
Online Resource 3, Table~S-3. Mistral and the
WikiGen column are the base model after guard rollback
(Section~\ref{sec:setup}). Discussed in Sections~\ref{sec:exp_trigger}
and~\ref{sec:mech_select}.}
\label{tab:mechanism}
\setlength{\tabcolsep}{4pt}
\fitwidth{tab:mechanism}{%
\begin{tabular}{lccccc}
\toprule
 & \multicolumn{4}{c}{\textbf{SciTech}} & \textbf{WikiGen} \\
 & \textbf{LLaMA} & \textbf{Qwen} & \textbf{Gemma} & \textbf{Mistral} & \textbf{LLaMA} \\
\midrule
Confidence ECE            & 0.686 & 0.522 & 0.537 & 0.722 & 0.493 \\
corr($p_{\mathrm{FT}}$, correct) & $+0.120$ & $+0.190$ & $+0.143$ & $+0.129$ & $+0.322$ \\
Selected-batch accuracy   & 0.000 & 0.000 & 0.031 & 0.000 & 0.000 \\
Random-batch accuracy     & 0.063 & 0.000 & 0.031 & 0.063 & 0.031 \\
Cycle-1/3 batch overlap   & 100\% & 100\% & 100\% & 100\% & 100\% \\
\bottomrule
\end{tabular}%
}
\end{table}

Table~\ref{tab:mechanism} reports the pre-specified diagnostics; Online Resource 3,
Table~S-6 extends them. Rank-based
metrics support the comparisons; the probability-scale
numbers describe the raw signal. Across the five combinations the
sequence-likelihood confidence carries an
expected calibration error of 0.49--0.72 and an AUROC against correctness of
0.65--0.81, with AURC 0.79--0.96; every interval is reported in Online Resource 3. Three
facts bound the interpretation. First, under the modest adaptation effects observed here, this is not a product of adaptation: the base
models, scored on the same queries, show similarly large ECE (0.48--0.72) and no
less discrimination (AUROC 0.73--0.81); adaptation moves calibration in both
directions; discrimination falls on LLaMA/SciTech (0.760 to 0.649 on the stored
record, to 0.602 regenerated under the corrected prompt template), holds on Qwen, and falls
on Gemma (0.726 to 0.676) --- a three-combination comparison for the intrinsic signals.
Second, it
is not specific to this operationalization: minimum token probability, mean token
entropy, and top-1/top-2 margin, computed from the same decoding pass, occupy the
same band (AUROC 0.57--0.82 on the domain corpora). Third, two of the five combinations are the base model after
guard rollback (Section~\ref{sec:setup}), so the profile describes the models as
deployed. What
the profile supports is a weak signal; what it does not support, on its own, is any
attribution of the failures below.

\subsection{Selection and attribution}
\label{sec:mech_select}

Under the registered containment diagnostic, selection over-represents its intended target: the
fraction of the selected batch that is retrieval-right and parametric-wrong is
0.81--1.00, meeting or exceeding the random distribution's 97.5th percentile on
every combination. Three facts stop us from interpreting this as evidence about the signal.
The target is nearly the same set as `retrieval right' --- the parametric mode
is wrong on 91--98\% of queries depending on the combination --- so context similarity alone
selects it as well as the full composite on three of five combinations
(Online Resource 3, Table~S-11) and the confidence term has almost
nothing to discriminate; the over-representation does not hold under a token-F1 criterion on
four of five combinations (Section~\ref{sec:mech_criterion}); and the converse direction's
target barely exists (0--9 queries per combination). These over-representation diagnostics therefore do not identify whether the confidence
term helps or hurts selection; the selector ablations below test that downstream contribution directly. The R-2 failure is a failure of a three-step transfer on
a frozen batch of 32; Online Resource 3, Table~S-2 also records that the random batch
outperformed the selected one on three of three LLaMA seeds and two of three Qwen
seeds, a difference we do not interpret at this budget.
We then did, on one combination. With a tenfold per-cycle budget (ten LoRA steps),
with per-cycle signals regenerated under the corrected prompt template,
LLaMA/SciTech distillation reached closed-book F1
0.078 --- below the undistilled 0.1005 and below a budget-matched random batch at
0.086 ($P > \text{off} = 0.000$, $P > \text{random} = 0.009$) --- and the capability
guard rolled it back at the second cycle (peak degradation 0.130; Online Resource 3,
Table~S-10).\footnote{The
per-cycle selected batches were not persisted, so the overlap between the
recomputed cycle-2 selection and cycle 1 is not recorded.} A larger budget did not
make the mechanism succeed; it degraded the model on both axes, and the random batch
degraded it too. On this combination, increasing the transfer budget tenfold does not restore an R-2 pass: both
the selected and the budget-matched random batch reduce closed-book F1. This
rules out the registered three-step budget as the sole explanation for this
combination, not for the mechanism generally. What remains unablated is the
objective itself.

The attribution we did not claim above can now be tested on the three genuinely
adapted combinations. On each, we ran the registered transfer --- same objective, same
three-step budget, same seeds --- from batches chosen by six selectors: the full
composite, the composite without the confidence term, the confidence term alone,
context similarity alone, disagreement alone, and a random batch (Online Resource 3, Sections~S7
and~S8, Tables~S-20 and~S-21; protocols committed before the runs; the
LLaMA combination under the corrected prompt template). Three combinations give three results, none providing evidence
supporting the confidence term. On LLaMA, removing the term never lowers
the outcome and on one seed raises it; the term alone does no better than
random. On Qwen, removing the term leaves the outcome inside the bootstrap width
on every seed and the term alone matches random --- 0.115--0.116 against
0.114--0.116; disagreement alone exceeds the random control (R-2) where the full
composite did not.
On Gemma, removing the confidence term raises the selector outcome relative to the
full composite (0.1026 / 0.1019 / 0.1031 against 0.0965 / 0.0974 / 0.0977, seed by seed; the
difference excludes zero on seed 42 by query, not by document cluster), and the composite
without the term passes both registered criteria (R-1 and R-2) whereas the full composite
passes neither; context similarity alone passes both as well. Confidence alone, by contrast,
falls below the random batch on three of three seeds (0.0910 / 0.0910 / 0.0888 against
0.0987 / 0.0964 / 0.0967). The registered verdicts stand as recorded --- they judge
the registered selector --- but the attribution we did not claim in
Section~\ref{sec:mech_select} resolves, on the combinations where it could be tested:
adding the confidence term yields no statistically detectable downstream benefit
on any of nine seeds. On Gemma, including the confidence term changes the
registered selector from one that passes R-1 and R-2 to one that fails both. We
state the
qualifications that bound this: three combinations, one prompt template, a three-step budget, and six
selectors compared under a pre-specified plan whose multiplicity we do not
correct for.

\subsection{The routing failure is a ceiling imposed by the setting}
\label{sec:mech_route}

The routing policy fails E-2 and E-4 because there is almost nothing to gain: retrieval
alone recovers 99.3--99.8\% of the best-case combined accuracy, only 5--14 of 2{,}860 queries are
answered parametrically and not by retrieval, and the gate chosen with hindsight
reduces to always-retrieval on every family (Section~\ref{sec:ablation}). On
the retrieval-failure subsets the possible gain is at most 0.011
(Section~\ref{sec:routing}).
Fixing the rank
calibration out of sample does not change this (Section~\ref{sec:routing}). The
registered
KBI offset improves the policy against $\gamma = 0$ but does not identify the
diagnostic's informativeness, both corpora lying in the same low-KBI band with the
offset's polarity opposite to its registered rationale.

\subsection{What recalibration can and cannot do}
\label{sec:mech_recal}

Scalar monotone recalibration --- Platt or temperature scaling applied to the
confidence score itself --- preserves ranks, and every selection rule in this paper
consumes the score only through ranks; the invariance is exact and, as an
implementation check, we confirm it (maximum difference 0 on all four families).
(The fitted scalar temperature saturates the search bound on every family: at
correct-answer rates of 2--6\%, the holdout likelihood is maximized by flattening
the signal toward a constant --- there is little to recover.)
This is a narrow claim and we state it narrowly: it says nothing about
recalibration applied to token logits, which reorders queries. We measured that
too: rescoring each generated sequence under token temperatures $T \in \{0.5, 2,
5\}$ changes the ranking (Spearman 0.65--0.97 against $T{=}1$) but improves AUROC
on no combination at $T{=}0.5$ and on five and four of thirteen combinations at $T{=}2$ and $5$
respectively, with selective accuracy following the same pattern
(Online Resource 3, Table~S-7). Temperature does not consistently recover the signal because there is little
to recover.

\subsection{Full fine-tuning does not improve the signal under the tested configuration}
\label{sec:mech_fullft}

Every adapted model above was trained with LoRA, which
leaves a last objection: the signal might fail only because
adapter training under-trains the model --- full
fine-tuning would produce a model that knows, and knows
that it knows. We test this with a control whose protocol,
constants, and reporting plan were committed before
execution, including a pledge to include the outcome
regardless of direction: full-parameter fine-tuning at the
same 1{,}000-step budget and effective batch, learning rate
fixed in advance at $2\times10^{-5}$ with no sweep,
capability guard off, three seeds per combination, on SciTech for
LLaMA, Qwen, and Mistral and on WikiGen for LLaMA.
Gemma-2-9B is the recorded exception: with a 256k-token
vocabulary, its optimizer residency alone (parameters,
gradients, and AdamW moments, $\approx$74\,GiB) exceeds
what a deployment-grade 80\,GB card can hold for full
fine-tuning in this precision configuration, and five
attempts failed identically at the first optimizer step; we
record it as an execution failure rather than substitute a
memory configuration the protocol prohibits.

Online Resource 3, Table~S-4 reports every seed.
Discrimination shows no detectable improvement on any combination --- per-seed
correlations sit below the LoRA record everywhere except
one Qwen seed at $+0.198$ against $+0.190$, within
bootstrap noise of a tie: they span $+0.072$--$0.081$ against the pre-fix LoRA
record of $+0.120$ (LLaMA, SciTech; $+0.088$ regenerated under the corrected
prompt template, Section~\ref{sec:setup}), $+0.225$--$0.239$
against $+0.322$ (LLaMA, WikiGen), $+0.024$--$0.045$
against $+0.129$ (Mistral), and $+0.173$--$0.198$ against
$+0.190$ (Qwen). Calibration error is mixed --- it improves
on two combinations and ties or worsens on the other two --- but
a large probability-scale mismatch remains (ECE above 0.50 on every
combination), and Section~\ref{sec:mech_recal} already
establishes that calibration is not the binding deficit. A
capability side-finding sharpens the point: at this budget,
full fine-tuning fails to exceed the recorded LoRA runs on
closed-book F1 on any combination (one-sided bootstrap
$P(\text{improve}) \leq 0.402$ everywhere; Mistral
collapses to 0.029--0.032 against 0.084), so the premise
that parameter-efficient adaptation under-trains these
models is itself unsupported in this setting. Three qualifications
bound the comparison, all fixed in the protocol: the
learning rate is a single conventional value, not tuned;
comparisons run against the as-recorded LoRA scalars
(guard-on runs) with unpaired bootstrap, since per-query
records of those runs were not retained; and the memory
configurations necessarily differ between methods. Within these bounds, full fine-tuning provides no evidence that LoRA is the
cause; the signal remains weak under this single pre-specified optimization
configuration.

\subsection{What depends on the correctness criterion}
\label{sec:mech_criterion}

The registered correctness criterion --- the answer contains the recorded gold span
--- is a lenient extractive standard, and it is lenient toward long answers. Rescoring
every diagnostic under token-F1 (at 0.3 and 0.5) and exact match (Online Resource 3,
Tables~S-13 and~S-14) shows what depends on it. The
setting does not: retrieval recovers 92.4--99.8\% of the best-case combined accuracy under every
criterion, and the parametric mode answers alone on at most 96 of 2{,}860
queries (3.4\%) under the most permissive. The signal's weakness does not: AUROC
falls from 0.65--0.81 to 0.55--0.72 under token-F1, and the parametric mode's
accuracy rises to 0.12--0.18 --- paraphrases the containment rule rejects ---
without the signal tracking it. The finer diagnostics do. Target over-representation,
significant on every combination under containment, loses significance on four of five
under token-F1; per-combination calibration numbers move with the criterion. One combination is
an outlier with a known cause: its answers were generated before a prompt-template
fix (Section~\ref{sec:setup}). LLaMA/SciTech's answers run to 45 words on
both modes against two-word gold spans, and containment credits them (retrieval
accuracy 0.81) while token-F1 does not (0.05); its registered diagnostics ---
including the 32-of-32 over-representation --- should be interpreted with that in mind, and answer
length correlates only weakly with containment elsewhere ($|\rho| \le 0.12$). Under
the corrected prompt template the combination's answer length is 7 words (median, against 45), the
parametric mode's containment and token-F1 accuracies 0.051 and 0.131 (against
0.059 and 0.016), and its diagnostics AUROC 0.602 and ECE 0.487 (against 0.649
and 0.686); the recorded combination's anomalies do not recur. We do
not choose a criterion after the fact: the registered one governs the
trigger verdicts (token-F1 on held-out test sets) and the routing pilot's verdict
(containment on the held-out query sets; Online Resource 3, Table~S-8 gives the split,
criterion, and resampling unit of every registered verdict), and every diagnostic
that depends on it is reported under the alternatives.
Finally, on the retrieval-failure subsets (gold chunk beyond rank 1 or rank 4),
the signal's discrimination is no better under token-F1 (AUROC 0.48--0.73) and
the parametric mode answers correctly 1--24 queries; the ceiling is not
an artifact of easy retrieval.

\section{Discussion}
\label{sec:discussion}

\subsection{What the negatives separate}
\label{sec:discussion_principle}

The instinctive interpretation of two failed confidence-based mechanisms --- the signal is
the cause --- is what this paper set out to test, and the data neither support
nor permit it (Sections~\ref{sec:mech_select}, \ref{sec:mech_route},
and~\ref{sec:mech_criterion}). What
the measurements do support is narrower and, we think, more useful: the signal is
weak, before and after adaptation, in every intrinsic form we measured; the
mechanism failures cannot generally be attributed to that weakness: the router
is ceiling-limited, while selector ablations settle the attribution against
the confidence term on Gemma.

\subsection{The rank-score calibration effect}
\label{sec:discussion_calibration}

The abstaining policy has lower ECE when ECE is computed on percentile-rank
scores, including at matched coverage. Because these scores are not
probabilities, we treat this as a rank-score alignment effect rather than
improved calibration. Whether calibrated abstention has value in a deployment
without an escalation path is a question for a setting in which a
probability-scale confidence is defined; this paper does not settle it.

\subsection{Future work}
\label{sec:discussion_future}

Three steps follow from what was not measured rather than from extending the
design: ablating the transfer objective
to localize the trigger's failure; a setting in which retrieval does not dominate,
to test whether the routing ceiling is specific to extractive QA; and a
pre-/post-adaptation comparison at larger scale, where base-model calibration may
differ. We do not propose a better trigger or router: the first would pursue an
unlocalized failure and the second a margin that does not exist here.

\subsection{A recommendation for practitioners}
\label{sec:discussion_practice}

We offer one recommendation drawn from this setting. Before building or evaluating a confidence-based router, measure
the possible routing gain --- the gap between the best-case combination and the best single mode --- on the
deployment's own query distribution; when that gap is a fraction of a percentage point, as here (0.2--0.5 points), no
confidence signal can be worth its cost, and the evaluation should turn to settings where retrieval fails.

\section{Limitations}
\label{sec:discussion_limitations}

The following conditions bound every conclusion above.

\textbf{The setting was built to give retrieval the advantage, and every verdict is
conditioned on it.} Questions were generated, by construction, from the
chunks containing their answers, so retrieval that surfaces
the right chunk surfaces the answer. This is realistic for
enterprise-style, source-grounded extractive document QA and unrepresentative for synthesis
across documents, multi-hop reasoning, or knowledge absent
from any retrievable passage. The routing ceiling --- the
finding on which the structural explanation of the routing verdict rests
--- is a property of this setting, and we do not claim it
holds where retrieval is weaker. The signal-quality half
(probability-scale mismatch and weak discrimination relative to the retrieval
mode) is measured
independently of the ceiling, but its generality beyond
this setting is likewise untested. The retrieval-failure subsets are defined on the held-out query sets by
gold-chunk rank; the test-split hard-negative subsets do not intersect them
(Section~\ref{sec:routing}).

\textbf{Coverage is uneven, and each result names its
base.} The trigger verdict covers four families on SciTech
with a WikiGen scale-up for LLaMA; the mechanism
diagnostics cover four families on SciTech and LLaMA alone
on WikiGen; the routing pilot ran on one model and two
corpora, with one registered ablation not run; the
full-fine-tuning control covers three of the four families,
the fourth being infeasible on deployment-grade single-GPU
memory; the contrast
corpus (NaturalQuestions, $n = 37$ validation examples,
anecdotal at that size)
received no adaptation treatment. Findings are
stated at the coverage that supports them, and none should
be interpreted more widely.

\textbf{Two combinations are unadapted.} Mistral/SciTech and
LLaMA/WikiGen are the base model after guard rollback
(Section~\ref{sec:setup}); every conclusion involving them
describes the models as deployed, not the effect of
adaptation on those combinations.

\textbf{The transfer objective is unablated.} The trigger's
registered transfer budget is 32 queries, one LoRA step per cycle,
three cycles; the budget was ablated on one combination
(Section~\ref{sec:mech_select}, exploratory) and made the outcome
worse, so the R-2 negative is not a negative for this budget alone ---
the distillation objective itself remains untested. The selector is now
ablated on three combinations (Section~\ref{sec:mech_select}); the objective is not.

\textbf{Resampling units.} Every registered bootstrap resamples queries, and
queries share documents (sixteen per document on SciTech); re-running every
registered bootstrap with document-level cluster resampling changes no verdict
(Online Resource 3, Table~S-8).

\textbf{Diagnostics are criterion-sensitive.} The setting
characterization and the signal's weakness hold under token-F1 and
exact-match criteria; target over-representation and per-combination calibration
do not (Section~\ref{sec:mech_criterion}), and we report them under
each criterion rather than choose one after the fact.

\textbf{One combination is a long-answer outlier with a known cause.}
LLaMA/SciTech's retention answers were generated, before the template
correction, with a prompt template the Llama-3.1 tokenizer does not
terminate (Section~\ref{sec:setup}); they average 45 words against two-word gold
spans, so the containment criterion credits them and token-F1 does not,
and that combination's registered diagnostics, including its 32-of-32
over-representation, are subject to that qualification. Regenerated diagnostics under the corrected
prompt template are in Online Resource 3, Table~S-6(c).

\textbf{One encoder underlies retrieval and two of the
three signals.} A single embedding model performs
retrieval, supplies the context-similarity proxy $\phi$, and
embeds the answers behind the disagreement signal $d$. A
systematic weakness of that encoder would move retrieval
quality and both derived signals together, in a way this
study cannot separate.

\textbf{The context-similarity proxy is coarse.} $\phi$ measures
embedding similarity between an answer and its retrieved
context --- a proxy for whether the answer is grounded in the retrieved context, not factual
correctness, so a fluent answer that restates the context
incorrectly scores highly. A local judge model would avoid
this at the cost of introducing a separate judge-based
reliability signal, which is outside the present scope.

\textbf{The verdict concerns four intrinsic sequence-level
signals, at one scale.} Separately trained confidence
estimators --- probes
in the style of \citet{kadavath2022language}, or
feature-based non-monotone recalibrators --- introduce an
additional reference signal and are outside the question
posed here. A pre-adaptation comparison now exists
(Section~\ref{sec:mech_diag}); all models fall in the
7--9B class, and whether the signal's weakness persists at
substantially larger scale, where base-model calibration
may differ, is outside what these
experiments measure.

\section{Conclusion}
\label{sec:conclusion}

We asked whether intrinsic sequence-likelihood confidence can serve as a control
signal in this retrieval-dominated extractive-QA setting. Under pre-specified
criteria, the trigger fails on all four families and the routing pilot fails on
LLaMA. The intrinsic signals are weak relative to retrieval, and the three
genuine pre-/post-adaptation comparisons provide no evidence that adaptation
repairs them. This conclusion is limited to the modest adaptation effects observed here. The failures nevertheless differ: on Gemma, removing confidence
changes the selector from failing to passing both registered criteria, whereas
the router faces essentially no possible gain even in the best case. The useful negative is
therefore a separation of signal, mechanism, and setting, not a general verdict
on confidence.

\backmatter

\bmhead{Supplementary information}
The code-and-configuration snapshot (version 5) and the corpus, question-answering, retention and probe files described in
Online Resource 3, Section~S4 are provided as Online Resource 1 and 2, a code record and a data record; both are also
deposited as Zenodo records (code: \url{https://doi.org/10.5281/zenodo.22710121}; data: \url{https://doi.org/10.5281/zenodo.22721044}).
Online Resource 1 (\texttt{ESM\_1.zip}) is the code record: code, protocols, analysis scripts, and result records.
Online Resource 2 (\texttt{ESM\_2.zip}) is the data record: the corpus chunk file, the generated question-answering sets, the registered
held-out query sets, the probe and concept sets, and the human label sheet.
Online Resource 3 (\texttt{ESM\_3.pdf}) contains Sections S1--S8: KBI construction, routing-protocol and distillation-trigger
details, the reproducibility statement, and the replication and ablation analyses cited in the text.

\bmhead{Acknowledgements}
This research was supported by Korea Institute of Science and Technology Information (KISTI) (No. (KISTI)K26L3M1C1, (NTIS)2710087347). This work was supported using HPC resources and technical support provided by KISTI.

\section*{Statements and Declarations}
\begin{itemize}
\item \textbf{Funding.} This research was supported by Korea Institute of Science and Technology Information (KISTI) (No. (KISTI)K26L3M1C1, (NTIS)2710087347).
\item \textbf{Competing interests.} The authors declare no competing interests.
\item \textbf{Ethics approval and consent to participate.} Not applicable: the study uses public corpora and no human subjects; the human labelling in the companion study was performed by an author.
\item \textbf{Data availability.} The generated question-answering sets, the registered held-out query sets, the corpus chunk file, the probe and concept sets, and the human label sheet are contained in the data record of the snapshot (version 5, 45 files plus \texttt{TREE\_MD5.txt}, md5 \texttt{ddc6e5e58103aa1907fc23ebb4bb07bd}) provided as Online Resource 2 and deposited at Zenodo~\citep{zenodo_data}. SciTech full text and verbatim answer sentences are not redistributed; identifiers, spans, and md5 values are, and a script rebuilds them; short answer spans ($\le 5$ tokens) are retained as quotations. Fifty question--answer items whose question asks for a person's contact details or the identity or affiliation of a paper's authors are excluded and listed with reasons; the figures reported here were computed on the full sets. The corpora derive from public resources (Online Resource 3, Section~S4).
\item \textbf{Code availability.} The code, protocols, analysis scripts, and result records are in the code record of the same snapshot (version 5, 1{,}395 files plus \texttt{TREE\_MD5.txt}, md5 \texttt{f3e17729a294e867968ea1419aa95e72}), provided as Online Resource 1 and deposited at Zenodo~\citep{zenodo_code}.
\item \textbf{Author contributions.} Gunwoo Lee: Conceptualization, Methodology, Software, Writing -- original draft, Visualization; Changmin Sung: Methodology, Investigation, Validation, Writing -- review \& editing; Sang-Hwan Gwak: Methodology, Validation; InA Kim: Software, Validation, Data curation; Ji-Young Choi: Investigation, Data curation, Visualization; Kyong-Ha Lee: Supervision, Funding acquisition, Writing -- review \& editing.
\item \textbf{Use of AI tools.} Language models were used to assist with prose drafting and editing and with the development of verification scripts. All experiments, analyses, scientific judgments, and the final manuscript were reviewed and verified by the authors.
\end{itemize}

\bibliography{references}

\clearpage
\global\backmatterfalse
\setcounter{secnumdepth}{3}
\setcounter{section}{0}
\renewcommand{\theHsection}{S\arabic{section}}\renewcommand{\theHsubsection}{S\arabic{section}.\arabic{subsection}}\renewcommand{\theHtable}{S-\arabic{table}}\renewcommand{\theHfigure}{S-\arabic{figure}}\renewcommand{\theHequation}{S\arabic{equation}}\ifdefined\theHalgorithm\renewcommand{\theHalgorithm}{S-\arabic{algorithm}}\fi
\section*{Online Resource 3}
\renewcommand{\thesection}{S\arabic{section}}
\renewcommand{\thetable}{S-\arabic{table}}
\renewcommand{\thefigure}{S-\arabic{figure}}
\renewcommand{\thealgorithm}{S-\arabic{algorithm}}
\renewcommand{\theequation}{S\arabic{equation}}
\setcounter{table}{0}\setcounter{figure}{0}\setcounter{algorithm}{0}\setcounter{equation}{0}
\noindent Tables, figures, and algorithms in this document are numbered S-1, S-2, \ldots; sections S1, S2, \ldots. References to ``Section'', ``Table'', and ``Figure'' without the S prefix point to the main text.

\FloatBarrier
\section{KBI Construction}
\label{app:kbi}

The corpus-level KBI consumed by the router is computed
before any adaptation. An extractor model identifies
$K = 50$ concepts in the corpus, and each concept is posed
to the base model as an identical fixed-template
closed-book probe; the response is scored against the
concept's source passage by ROUGE-L F-measure and by
embedding cosine similarity under the deployment encoder.
The corpus-level KBI is the mean probe score, so higher KBI
means the base model already reproduces more of the corpus
content. The router consumes the ROUGE-L scalar (0.164 for
SciTech, 0.244 for WikiGen).

\FloatBarrier
\section{Routing Protocol: Design Detail}
\label{sup:routing}

This section gives the routing and abstention rules of
Section~7 of the main text in full. The four registered
acceptance criteria, the fixed constants, and the verdict
correction are stated in the main text and are not repeated
here.

\FloatBarrier
\subsection{Mode scores}
\label{sup:routing_scores}

Both mode scores are built from the per-query signals of
Section~3 of the main text, converted to
percentile ranks
within the evaluation set:
\begin{align}
  S_{\mathrm{RAG}}(q) &= \phi_{\mathrm{rank}}(q)
    \label{eq:ens_rag_score}\\[2pt]
  S_{\mathrm{FT}}(q)  &= p_{\mathrm{FT,rank}}(q)
    \;-\; \gamma\,\mathrm{KBI}(\mathcal{D})
    \label{eq:ens_ft_score}
\end{align}
with $\gamma = 2.0$. The corpus-level \kbs{} term is the
only place where the router consults the knowledge-coverage
diagnostic. Its role is a
prior: on a corpus the base model barely covers, the
parametric mode should need a higher query-level
confidence to be selected. Otherwise the query
is answered by the higher-scoring mode. This is the
rationale as registered; Section~7 records
that it uses KBI in the opposite sense from its measured
polarity.

\FloatBarrier
\subsection{Abstention gating}
\label{sup:routing_gating}

The router declines in two mutually exclusive
circumstances. It abstains on \emph{low confidence} when
the higher score falls below
$\theta_{\mathrm{abstain}} = 0.3$. It abstains on
\emph{conflict} when both modes are independently
confident yet disagree about which should answer:
formally, when $\phi_{\mathrm{rank}}(q) >
\theta_{\mathrm{high}}$ and $p_{\mathrm{FT,rank}}(q) >
\theta_{\mathrm{high}}$ with $\theta_{\mathrm{high}} = 0.7$,
\emph{and} the two routing scores are within
$\theta_{\mathrm{conflict}} = 0.15$ of each other.

The two conditions are tested in that order and cannot
both hold: the conflict test requires
$\phi_{\mathrm{rank}} > 0.7$, hence
$S_{\mathrm{RAG}} > 0.7 > \theta_{\mathrm{abstain}}$, so
the low-confidence branch is already false whenever the
conflict branch is reachable.

\FloatBarrier
\subsection{Why the gate and
the margin use different scales}
\label{sup:routing_twospace}

The gate is evaluated on \emph{raw} percentile ranks, the
margin on the \kbs-adjusted scores of
Eqs.~\eqref{eq:ens_rag_score}--\eqref{eq:ens_ft_score}. The
reason is arithmetic. $S_{\mathrm{FT}}$ is bounded above by
$1 - \gamma\,\mathrm{KBI}$, which for the corpora used here
is at most \textit{0.67}; gating on the adjusted score
would place the parametric mode permanently below
$\theta_{\mathrm{high}}$ and render the conflict branch
unreachable. Evaluating the gate on raw ranks keeps the
branch live while leaving the \kbs{} prior to act where it
is meant to, on the routing preference. This two-space
design was introduced in the v3 protocol specifically to
repair that dead branch. Section~7 of the
main text reports how far the repair is vindicated by the data.

\FloatBarrier
\section{AKD Formulation: Detail}
\label{sup:akd}\label{app:trigger}

This section gives the signal normalization, the composite
selection scores and the \kbs-adaptive rate cap of
Section~5 of the main text in full,
together with the cost decomposition. The disagreement
signal $d(q)$ remains in the main text as
Eq.~1; the distillation objective is
stated below as Eq.~\ref{eq:akd_loss_app}.

\FloatBarrier
\subsection{Rank normalization}
\label{sup:akd_rank}

The empirical ranges of the three raw signals differ
substantially---$d$ concentrates near zero under strong
embedding similarity, the embedding-based $\phi$ is bounded
away from zero, and $p_{\mathrm{FT}}$ ranges broadly over
$[0,1]$---so na\"ively summing them would let one signal
dominate the composite. Each signal is therefore normalized
to its percentile rank within the held-out query set:
\begin{equation}
  x_{\mathrm{rank}}(q) =
    \frac{\bigl|\{q' \in \mathcal{Q}_{\mathrm{ret}}
                : x(q') \leq x(q)\}\bigr|}
         {|\mathcal{Q}_{\mathrm{ret}}|}
  \label{eq:akd_rank_norm}
\end{equation}
for $x \in \{d, \phi, p_{\mathrm{FT}}\}$ (ties averaged),
producing $d_{\mathrm{rank}}, \phi_{\mathrm{rank}},
p_{\mathrm{FT,rank}} \in [0,1]$ each with uniform marginal.
This places all three signals on a common scale and
provides robustness to outliers.

\FloatBarrier
\subsection{Directional composite scores}
\label{sup:akd_scores}

For each retained query $q \in \mathcal{Q}_{\mathrm{ret}}$,
\akd{} combines the three ranked signals into two
directional composite scores:
\begin{align}
  S_{\mathrm{R}\to\mathrm{F}}(q)
    &= w_1\,d_{\mathrm{rank}}(q)
     + w_2\,\phi_{\mathrm{rank}}(q)
     + w_3\,\bigl(1 - p_{\mathrm{FT,rank}}(q)\bigr)
    \label{eq:akd_score_r2f}\\[2pt]
  S_{\mathrm{F}\to\mathrm{R}}(q)
    &= w_1\,d_{\mathrm{rank}}(q)
     + w_2\,\bigl(1 - \phi_{\mathrm{rank}}(q)\bigr)
     + w_3\,p_{\mathrm{FT,rank}}(q)
    \label{eq:akd_score_f2r}
\end{align}
with default weights $(w_1, w_2, w_3) = (1,1,1)$,
representing equal contribution of the three ranked
signals.

The two scores share the disagreement term
$d_{\mathrm{rank}}$: a distillation candidate must exhibit
mode disagreement in \emph{either} direction. The
reliability signals enter with opposite polarity across the
two directions. $S_{\mathrm{R}\to\mathrm{F}}$ prefers
queries where the RAG answer is grounded in reliable
retrieval and the FT answer is under-confident (reliable
teacher, uncertain student), while
$S_{\mathrm{F}\to\mathrm{R}}$ prefers queries where the FT
answer is high-confidence and the RAG answer is poorly
grounded (confident teacher, unreliable student).

Selection is by top-$N$ retrieval on each composite score
rather than by threshold gating. This budget-aware
formulation guarantees non-degenerate distillation batches
whenever $N \leq |\mathcal{Q}_{\mathrm{ret}}|$, avoiding
the empty-selection case that arises when trigger conditions
are strict conjunctions of tail percentiles.

The percentile-gate path
remains in the implementation
(\texttt{akd/trigger.py}) but is not the path
exercised by the scale-up experiments reported in
Section~6.

\FloatBarrier
\subsection{Distillation objective}
\label{sup:akd_obj}
For the RAG$\to$FT direction we treat the RAG mode's
context-conditioned response as a soft teacher signal.
Writing $P_{\mathrm{FT}}(\cdot \mid q;
\theta_{\mathrm{FT}})$ for the current FT model's response
distribution and $P_{\mathrm{FT}}(\cdot \mid q,
c_{\mathrm{RAG}}; \theta_{\mathrm{FT}}^{\mathrm{frozen}})$
for the same model conditioned on the RAG retrieval with
weights frozen at the start of the step, the loss is a
task--distillation combination:

\begin{equation}
  \begin{aligned}
  \mathcal{L}_{\mathrm{AKD}}^{R \to F}
  ={}& \alpha_{\mathrm{AKD}}\,
    \mathcal{L}_{\mathrm{task}}\bigl(
      \theta_{\mathrm{FT}};\, q,\, a_{\mathrm{RAG}}
    \bigr) \\
  &{}+ (1 - \alpha_{\mathrm{AKD}})\,
    D_{\mathrm{KL}}\!\Bigl(
      P_{\mathrm{FT}}(\cdot \mid q;\,
                      \theta_{\mathrm{FT}}) \\
  &\qquad\quad {}\Big\|\;
      P_{\mathrm{FT}}(\cdot \mid q, c_{\mathrm{RAG}};\,
                      \theta_{\mathrm{FT}}^{\mathrm{frozen}})
    \Bigr)
  \end{aligned}
  \label{eq:akd_loss_app}
\end{equation}

with $\alpha_{\mathrm{AKD}} = 0.5$,
$\mathcal{L}_{\mathrm{task}}$ the masked cross-entropy on
the retrieval-mode answer $a_{\mathrm{RAG}}$, and the
divergence at temperature
$T = 1$. The first term preserves domain task fit; the
second pulls the FT model's parametric response toward what
the same model produces with retrieved evidence available.
Teacher logits are detached, so gradients flow only through
the student branch, and updates use the same LoRA
adapters~\cite{hu2022lora} as the original fine-tuning
stage.

\FloatBarrier
\subsection{\kbs-adaptive rate cap}
\label{sup:akd_rate}\label{sec:akd_rate}

To bound the cost of \akd{} and concentrate distillation
effort where it is most likely to pay off, the per-cycle
trigger rate is capped as a function of the corpus \kbs{}
measured before adaptation:
\begin{equation}
  r_{\mathrm{AKD}}(\mathcal{D})
  =
  \begin{cases}
    r_{\mathrm{high}}
      & \text{if } \mathrm{KBI}(\mathcal{D}, M)
                    < \theta_L \\[2pt]
    r_{\mathrm{mid}}
      & \text{if } \theta_L \leq
                   \mathrm{KBI}(\mathcal{D}, M) \leq \theta_H \\[2pt]
    r_{\mathrm{low}}
      & \text{if } \mathrm{KBI}(\mathcal{D}, M)
                    > \theta_H
  \end{cases}
  \label{eq:akd_rate}
\end{equation}
with default caps $(r_{\mathrm{high}}, r_{\mathrm{mid}},
r_{\mathrm{low}}) = (0.30, 0.15, 0.05)$. The mid band
provides a conservative default between the two outer
bands. For
low-\kbs{} corpora the base model has substantial scope to
absorb corpus knowledge, so up to 30\% of retained queries
may be used per cycle; for high-\kbs{} corpora the base
model is already strong on the domain, so distillation runs
sparingly. The cap applies equally to both directions: only
the rate, not the direction, is \kbs-dependent. The
thresholds $\theta_L = 0.35$ and $\theta_H = 0.65$ reuse
the values calibrated for corpus segmentation, so \akd{}
introduces no additional threshold-tuning burden.

The cap enters the selection rule through $N =
\min\bigl(B_{\mathrm{AKD}}, \lceil r_{\mathrm{AKD}}
|\mathcal{Q}_{\mathrm{ret}}| \rceil\bigr)$, evaluated
identically for both directions: top-$N$ selection by
composite score subsumes both budget matching and priority
tie-breaking, so no separate deferral rule is required.
Section~5 of the main text records
that the tiering is a
design-time provision rather than an empirically exercised
mechanism in this work.

\FloatBarrier
\subsection{Cost decomposition}
\label{sup:akd_cost}

\akd{} cost decomposes into (i) a one-time retention build
cost per query in $\mathcal{Q}_{\mathrm{ret}}$, comprising
open-book and closed-book greedy generation (each at most
$L_{\max} = 64$ decoded tokens from the same adapted model,
differing only in the prompt) and two bge-m3 embedding
passes---one for retrieval-side chunk selection, one for
the context-similarity proxy $\phi(q)$---amortized across all
cycles, and (ii) $C$ LoRA update steps per direction during
refinement. The FT confidence $p_{\mathrm{FT}}(q)$ is read
off the generation scores and incurs no separate cost, and
the answer embeddings for $d(q)$ are computed at cycle-loop
entry from the stored answer texts rather than during the
retention build.

Because the composite scores and the top-$N$ selection are
computed once before the cycle loop and reused, no
additional forward passes are incurred per cycle beyond the
LoRA update and a check of the general-capability benchmark set
(Section~\ref{app:repro}); the per-cycle
cost is dominated by the LoRA gradient step over a batch of
size $N$. The per-cycle scoring cost thus collapses to
zero, trading exploration across cycles for deterministic,
replicable selection --- a design consequence we analyze in
Section~9. The one-time computation of rank-normalized
signals requires sorting three
$|\mathcal{Q}_{\mathrm{ret}}|$-length vectors
($\mathcal{O}(|\mathcal{Q}_{\mathrm{ret}}|
\log|\mathcal{Q}_{\mathrm{ret}}|)$), and top-$N$ selection
uses a heap of size $N$
($\mathcal{O}(|\mathcal{Q}_{\mathrm{ret}}| \log N)$); both
are absorbed into the scoring pass.

\FloatBarrier
\subsection{\akd{} Refinement Loop}
\label{sup:alg_akd}

Algorithm~\ref{alg:akd} summarizes the full \akd{} refinement
loop, which runs as a post-adaptation stage after the
initial adaptation, run under the capability-regression
guard, completes (Section~5). The
composite selection scores and the rate cap it invokes are
given in Section~\ref{sup:akd}; the disagreement signal
$d(q)$ is Eq.~1 of the main text and the
distillation objective is Eq.~\ref{eq:akd_loss_app}.

\begin{algorithm}[htbp]
\caption{\akd{} Refinement Loop}
\label{alg:akd}
\footnotesize
\begin{algorithmic}[1]
\REQUIRE Post-adaptation FT model $M_{\mathrm{FT}}$,
         RAG pipeline $\mathcal{R}$,
         held-out query set $\mathcal{Q}_{\mathrm{ret}}$
         with precomputed $\phi(q)$, $p_{\mathrm{FT}}(q)$,
         corpus $\mathcal{D}$,
         \kbs{} value measured before adaptation,
         max cycles $C$
\ENSURE  Refined FT model $M_{\mathrm{FT}}^\star$ and
         refined retrieval preference set $\mathcal{P}^\star$
\STATE Compute rate cap $r_{\mathrm{AKD}}$ via
       Eq.~\eqref{eq:akd_rate}
\STATE Load precomputed signals $\phi(q)$,
       $p_{\mathrm{FT}}(q)$ for all
       $q \in \mathcal{Q}_{\mathrm{ret}}$; compute the
       answer-embedding disagreement $d(q)$
       (Eq.~1 of the main text) online
\STATE Normalize each signal to percentile rank
       (Eq.~\eqref{eq:akd_rank_norm}), obtaining
       $(d_{\mathrm{rank}}, \phi_{\mathrm{rank}},
         p_{\mathrm{FT,rank}})$
\STATE Compute composite scores
       $S_{\mathrm{R}\to\mathrm{F}}(q)$ and
       $S_{\mathrm{F}\to\mathrm{R}}(q)$
       (Eqs.~\eqref{eq:akd_score_r2f}--\eqref{eq:akd_score_f2r})
\STATE Set selection budget
       $N \gets \min\!\bigl(B_{\mathrm{AKD}},\;
       \lceil r_{\mathrm{AKD}} \cdot
              |\mathcal{Q}_{\mathrm{ret}}| \rceil\bigr)$
       for each direction
\STATE \COMMENT{Composite scores and selection budget are
       computed once, before the cycle loop; they are not
       recomputed as the adapter updates.}
\FOR{cycle $= 1$ \TO $C$}
  \STATE $\mathcal{B}_{R \to F} \gets
         \operatorname{top\text{-}N}\!\bigl(
         \mathcal{Q}_{\mathrm{ret}},\;
         S_{\mathrm{R}\to\mathrm{F}}\bigr)$
  \STATE $\mathcal{B}_{F \to R} \gets
         \operatorname{top\text{-}N}\!\bigl(
         \mathcal{Q}_{\mathrm{ret}},\;
         S_{\mathrm{F}\to\mathrm{R}}\bigr)$
  \STATE Update $M_{\mathrm{FT}}$ via one LoRA step
         on $\mathcal{B}_{R \to F}$ minimizing
         Eq.~\ref{eq:akd_loss_app}
  \STATE Add $\mathcal{B}_{F \to R}$ answers to
         retrieval preference set $\mathcal{P}$
  \STATE Run the general-capability benchmark set
         (Section~\ref{app:repro}) on updated
         $M_{\mathrm{FT}}$; if degradation exceeds
         $\beta$, roll back and terminate loop
\ENDFOR
\STATE $M_{\mathrm{FT}}^\star \leftarrow M_{\mathrm{FT}}$;\;
       $\mathcal{P}^\star \leftarrow \mathcal{P}$
\end{algorithmic}
\end{algorithm}

\begin{table}[htbp]
\centering
\footnotesize
\caption{Component ablation across four base models. Accuracy
on the held-out query set. The $p_{\mathrm{FT}}$ and $\phi$ gates
are evaluated at best-case thresholds. Plotted as Figure~1; discussed in
Section~8.}
\label{tab:component_ablation}
\setlength{\tabcolsep}{5pt}
\begin{tabular}{lcccccc}
\toprule
\textbf{Base model} & \textbf{FT} & \textbf{RAG}
  & \textbf{Best-case} & \textbf{$p_{\mathrm{FT}}$}
  & \textbf{$\phi$} & \textbf{Rand.} \\
 & \textbf{only} & \textbf{only} & \textbf{comb.}
  & \textbf{gate} & \textbf{gate} & \textbf{gate} \\
\midrule
LLaMA-3.1-8B & 0.073 & 0.809 & 0.811 & 0.809 & 0.809 & 0.756 \\
Qwen2.5-7B   & 0.059 & 0.849 & 0.852 & 0.849 & 0.849 & 0.807 \\
Gemma-2-9B   & 0.081 & 0.848 & 0.851 & 0.848 & 0.848 & 0.788 \\
Mistral-7B   & 0.058 & 0.735 & 0.740 & 0.735 & 0.735 & 0.699 \\
\bottomrule
\end{tabular}
\end{table}

\begin{table}[htbp]
\centering
\footnotesize
\caption{Seed-level outcomes for the retrieval-to-parametric
direction. $P_{>\mathrm{off}}$ and $P_{>\mathrm{rand}}$ are
bootstrap probabilities of exceeding the \akd-off and
random-batch controls respectively; the pre-specified decision
threshold is $0.7$. Plotted as Figure~\ref{fig:akd};
discussed in Section~6.}
\label{tab:akd_seed}
\setlength{\tabcolsep}{5pt}
\begin{tabular}{llccccc}
\toprule
\textbf{Model} & \textbf{Seed} & \textbf{Off}
  & \textbf{R$\to$F} & \textbf{Rand.}
  & \textbf{$P_{>\mathrm{off}}$}
  & \textbf{$P_{>\mathrm{rand}}$} \\
\midrule
\multirow{3}{*}{LLaMA-3.1-8B}
  & 42 & 0.1005 & 0.0959 & 0.1023 & 0.139 & 0.011 \\
  & 43 & 0.1005 & 0.0961 & 0.1040 & 0.165 & 0.003 \\
  & 44 & 0.1005 & 0.0959 & 0.1024 & 0.136 & 0.030 \\
\midrule
\multirow{3}{*}{Qwen2.5-7B}
  & 42 & 0.1116 & 0.1142 & 0.1189 & 0.848 & 0.048 \\
  & 43 & 0.1116 & 0.1159 & 0.1145 & 0.946 & 0.677 \\
  & 44 & 0.1116 & 0.1145 & 0.1178 & 0.874 & 0.149 \\
\midrule
\multirow{3}{*}{Gemma-2-9B}
  & 42 & 0.0990 & 0.0965 & 0.0987 & 0.259 & 0.267 \\
  & 43 & 0.0990 & 0.0974 & 0.0964 & 0.312 & 0.629 \\
  & 44 & 0.0990 & 0.0977 & 0.0967 & 0.337 & 0.612 \\
\midrule
\multirow{3}{*}{Mistral-7B}
  & 42 & 0.0843 & 0.0855 & 0.0842 & 0.632 & 0.637 \\
  & 43 & 0.0843 & 0.0865 & 0.0811 & 0.726 & 0.919 \\
  & 44 & 0.0843 & 0.0860 & 0.0850 & 0.698 & 0.610 \\
\bottomrule
\end{tabular}
\end{table}

\begin{figure}[!htb]
  \centering
  \includegraphics[width=\linewidth]{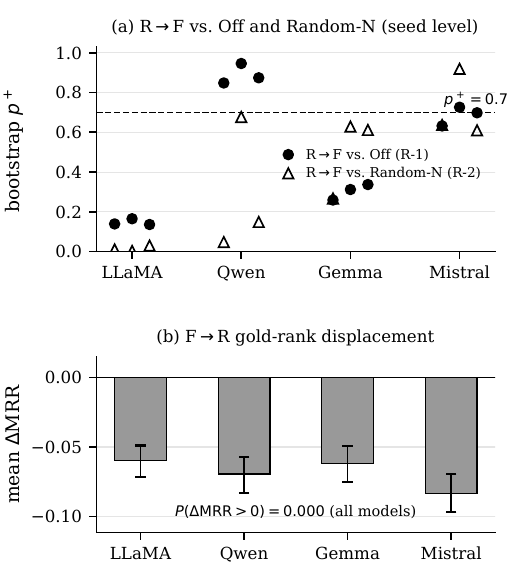}
  \caption{Seed-level distillation outcomes and gold-rank displacement.}
  \label{fig:akd}
\end{figure}
\FloatBarrier

\begin{table}[htbp]
\centering
\footnotesize
\caption{Parametric-to-retrieval displacement (panel~(a) of
the trigger diagnostics; discussed in
Section~9.1).}
\label{tab:displacement}
\setlength{\tabcolsep}{5pt}
\begin{tabular}{lccc}
\toprule
\textbf{Base model} & \textbf{Prefs.}
  & \textbf{Mean $\Delta$MRR} & \textbf{$P_{\Delta>0}$} \\
\midrule
LLaMA-3.1-8B & 31 & $-0.060$ & 0.000 \\
Qwen2.5-7B   & 31 & $-0.070$ & 0.000 \\
Gemma-2-9B   & 31 & $-0.062$ & 0.000 \\
Mistral-7B   & 31 & $-0.083$ & 0.000 \\
\bottomrule
\end{tabular}
\end{table}

\begin{table}[htbp]
\centering
\scriptsize
\caption{Full-fine-tuning control (B-2), every seed reported.
LoRA records are as-recorded guard-on runs (scale-up off
condition for F1; retention diagnostics for ECE/corr); for
Mistral/SciTech and LLaMA/WikiGen the record is a
base-model record (guard rollback;
Section~4).
$P(\text{improve})$ is a one-sided \emph{unpaired} bootstrap
against the recorded LoRA scalar (per-query records of the
LoRA runs were not retained). Discussed in
Section~9.5. The LLaMA/SciTech LoRA record predates the
template fix; the regenerated record under the corrected prompt template is corr
$+0.088$, ECE 0.487.}
\label{tab:b2}
\setlength{\tabcolsep}{4pt}
\fitwidth{tab:b2}{%
\begin{tabular}{llcccccc}
\toprule
\textbf{Model} & \textbf{Corpus}
  & \textbf{full-FT F1 (s42/43/44)} & \textbf{LoRA F1}
  & \textbf{$P(\text{improve})$} & \textbf{full-FT ECE}
  & \textbf{full-FT corr} & \textbf{LoRA ECE/corr} \\
\midrule
LLaMA-3.1-8B & SciTech & 0.0875/0.0962/0.0929 & 0.1005 & 0.002--0.211 & 0.557--0.592 & $+0.072$--$0.081$ & 0.686/$+0.120$ \\
LLaMA-3.1-8B & WikiGen & 0.1694/0.1632/0.1609 & 0.1758 & 0.019--0.190 & 0.503--0.521 & $+0.225$--$0.239$ & 0.493/$+0.322$ \\
Qwen2.5-7B   & SciTech & 0.1101/0.1093/0.1093 & 0.1116 & 0.348--0.402 & 0.559--0.565 & $+0.173$--$0.198$ & 0.522/$+0.190$ \\
Mistral-7B   & SciTech & 0.0287/0.0300/0.0315 & 0.0843 & 0.000--0.000 & 0.572--0.587 & $+0.024$--$0.045$ & 0.722/$+0.129$ \\
Gemma-2-9B   & SciTech & \multicolumn{6}{l}{infeasible --- OOM at first optimizer step ($\approx$74\,GiB optimizer residency); see text} \\
\bottomrule
\end{tabular}%
}
\end{table}

\begin{table}[htbp]
\centering
\footnotesize
\caption{Target over-representation for the trigger's target (retrieval right
$\wedge$ parametric wrong). Selected batches are the deterministic top-$N$
re-computation of the recorded selection; random-$N$ is 10{,}000 draws of 32
without replacement (seed 42); $P(X \ge k)$ is the exact hypergeometric tail.
Discussed in Section~9.2.}
\label{tab:enrichment}
\setlength{\tabcolsep}{5pt}
\fitwidth{tab:enrichment}{%
\begin{tabular}{lcccc}
\toprule
\textbf{Model / corpus} & \textbf{Population target rate}
  & \textbf{Selected $k/32$} & \textbf{Random median [2.5, 97.5\%]}
  & \textbf{$P(X \ge k)$} \\
\midrule
LLaMA/SciTech & 894/1196 (0.747) & 32/32 (1.000) & 24 [19, 28] & $7.8\times10^{-5}$ \\
Qwen/SciTech & 845/1196 (0.707) & 28/32 (0.875) & 23 [18, 27] & 0.021 \\
Gemma/SciTech & 814/1196 (0.681) & 28/32 (0.875) & 22 [17, 27] & 0.010 \\
Mistral/SciTech & 783/1196 (0.655) & 26/32 (0.812) & 21 [16, 26] & 0.039 \\
LLaMA/WikiGen & 1275/1759 (0.725) & 28/32 (0.875) & 23 [18, 28] & 0.036 \\
\bottomrule
\end{tabular}
}
\end{table}

\begin{table}[htbp]
\centering
\scriptsize
\caption{The signal, profiled. Panel (a): six metrics of the recorded (deployed)
sequence-likelihood signal per combination, with 95\% bootstrap CIs ($B = 10{,}000$,
seed 42). Panel (b): the same signal from a regeneration pass of the \emph{base}
models on the same queries. Panel (c): AUROC of the four intrinsic signals from
the regeneration pass (entropy reported oriented, as $1 - \mathrm{AUROC}$ of raw
entropy, so higher = more confident); NQ rows are the $n = 37$ contrast set. Combinations
labeled \emph{adapted} for Mistral/SciTech and LLaMA/WikiGen are the base model
after guard rollback (Section~4). The adapted LLaMA/SciTech row of
panel (c) was regenerated under the corrected prompt template
(Section~4; corr $+0.088$, ECE 0.487, token-temperature AUROC
0.589/0.602/0.598/0.599 at $T = 0.5/1/2/5$); the stored record (AUROC 0.649, corr
$+0.120$, ECE 0.686) is the pre-fix artifact. Discussed in
Section~9.1.}
\label{tab:signal_profile}
\setlength{\tabcolsep}{3pt}
\fitwidth{tab:signal_profile}{%
\begin{tabular}{lccccc}
\toprule
 & \textbf{LLaMA/Sci} & \textbf{Qwen/Sci} & \textbf{Gemma/Sci} & \textbf{Mistral/Sci} & \textbf{LLaMA/Wiki} \\
\midrule
\multicolumn{6}{l}{\emph{(a) deployed signal (stored retention record)}}\\
AUROC & 0.649 [0.580, 0.716] & 0.736 [0.673, 0.794] & 0.677 [0.599, 0.754] & 0.768 [0.665, 0.860] & 0.808 [0.771, 0.842] \\
AUPRC & 0.135 [0.086, 0.216] & 0.179 [0.107, 0.283] & 0.175 [0.091, 0.274] & 0.081 [0.037, 0.167] & 0.404 [0.327, 0.481] \\
AURC & 0.903 [0.874, 0.930] & 0.899 [0.869, 0.927] & 0.915 [0.888, 0.941] & 0.959 [0.938, 0.977] & 0.789 [0.759, 0.820] \\
E-AURC & 0.130 [0.104, 0.154] & 0.101 [0.079, 0.124] & 0.100 [0.077, 0.124] & 0.053 [0.036, 0.071] & 0.101 [0.085, 0.118] \\
Brier & 0.531 [0.520, 0.541] & 0.333 [0.323, 0.344] & 0.341 [0.332, 0.350] & 0.549 [0.538, 0.558] & 0.325 [0.315, 0.334] \\
NLL & 1.365 [1.338, 1.393] & 0.887 [0.863, 0.912] & 0.896 [0.876, 0.917] & 1.436 [1.404, 1.468] & 0.871 [0.849, 0.894] \\
\midrule
\multicolumn{6}{l}{\emph{(b) base model, regeneration pass (same queries)}}\\
ECE (10-bin) & 0.480 & 0.719 & 0.654 & 0.722 & 0.493 \\
corr & $+0.116$ & $+0.089$ & $+0.112$ & $+0.129$ & $+0.322$ \\
AUROC & 0.760 [0.647, 0.851] & 0.726 [0.592, 0.844] & 0.726 [0.612, 0.827] & 0.768 [0.664, 0.863] & 0.808 [0.771, 0.842] \\
AURC & 0.969 & 0.971 & 0.957 & 0.959 & 0.789 \\
\bottomrule
\end{tabular}%
}
\vspace{4pt}

\fitwidth{tab:signal_profile}{%
\begin{tabular}{lcccc}
\toprule
\emph{(c) intrinsic AUROC} & \textbf{$p_{\mathrm{seq}}$} & \textbf{min prob} & \textbf{entropy (or.)} & \textbf{margin} \\
\midrule
base LLaMA/SciTech & 0.760 & 0.675 & 0.781 & 0.704 \\
base Qwen/SciTech & 0.726 & 0.699 & 0.753 & 0.711 \\
base Gemma/SciTech & 0.726 & 0.734 & 0.759 & 0.702 \\
base Mistral/SciTech & 0.768 & 0.638 & 0.773 & 0.772 \\
base LLaMA/WikiGen & 0.808 & 0.818 & 0.814 & 0.757 \\
adapted Qwen/SciTech & 0.736 & 0.627 & 0.724 & 0.713 \\
adapted Gemma/SciTech & 0.676 & 0.636 & 0.658 & 0.679 \\
adapted LLaMA/SciTech (regenerated, corrected prompt template) & 0.602 & 0.573 & 0.598 & 0.577 \\
adapted Mistral/SciTech (= base) & 0.768 & 0.638 & 0.773 & 0.772 \\
adapted LLaMA/WikiGen (= base) & 0.808 & 0.818 & 0.814 & 0.757 \\
base LLaMA/NQ & 0.813 & 0.794 & 0.829 & 0.758 \\
base Qwen/NQ & 0.640 & 0.613 & 0.656 & 0.672 \\
base Gemma/NQ & 0.681 & 0.600 & 0.719 & 0.722 \\
base Mistral/NQ & 0.742 & 0.801 & 0.742 & 0.699 \\
\bottomrule
\end{tabular}%
}
\end{table}

\begin{table}[htbp]
\centering
\scriptsize
\caption{Token-temperature grid: AUROC of the sequence log-likelihood rescored
under softmax(logit$/T$), per combination. Against $T{=}1$, AUROC improves on 0/13 combinations
at $T{=}0.5$, 5/13 at $T{=}2$, and 4/13 at $T{=}5$; the $T{\ne}1$ orderings differ
from $T{=}1$ on every combination (Spearman 0.65--0.97); selective accuracy at coverage
0.5--0.9 follows the same pattern. Discussed in Section~9.4.}
\label{tab:tgrid}
\setlength{\tabcolsep}{5pt}
\begin{tabular}{lcccc}
\toprule
\textbf{Model / corpus} & \textbf{$T{=}0.5$} & \textbf{$T{=}1$} & \textbf{$T{=}2$} & \textbf{$T{=}5$} \\
\midrule
base LLaMA/SciTech & 0.678 & 0.760 & 0.755 & 0.738 \\
base Qwen/SciTech & 0.684 & 0.726 & 0.737 & 0.780 \\
base Gemma/SciTech & 0.659 & 0.726 & 0.744 & 0.718 \\
base Mistral/SciTech & 0.728 & 0.768 & 0.765 & 0.764 \\
base LLaMA/WikiGen & 0.767 & 0.808 & 0.768 & 0.739 \\
adapted Qwen/SciTech & 0.733 & 0.736 & 0.748 & 0.762 \\
adapted Gemma/SciTech & 0.668 & 0.676 & 0.648 & 0.646 \\
adapted Mistral/SciTech (= base) & 0.728 & 0.768 & 0.765 & 0.764 \\
adapted LLaMA/WikiGen (= base) & 0.767 & 0.808 & 0.768 & 0.739 \\
base LLaMA/NQ & 0.742 & 0.813 & 0.813 & 0.806 \\
base Qwen/NQ & 0.629 & 0.640 & 0.661 & 0.656 \\
base Gemma/NQ & 0.652 & 0.681 & 0.644 & 0.637 \\
base Mistral/NQ & 0.731 & 0.742 & 0.801 & 0.763 \\
\bottomrule
\end{tabular}
\end{table}

\FloatBarrier
\section{Reproducibility statement}
\label{app:repro}

All experiments ran on a pinned software stack --- Python
3.10, torch 2.5.1, transformers 4.47.1, peft 0.19.0,
accelerate 1.13.0 --- on A100 80\,GB GPUs.

General capability is the \cas{} Quick Profile---MMLU (57 subjects
$\times$ 100 items $= 5{,}700$), ARC-Challenge (100), KMMLU (45 subjects
$\times$ 50 $= 2{,}250$); identical items at every evaluation, deterministic
within a session (zero item disagreement on re-evaluation), while across sessions
the baseline moved by 4, 1, and 7 items on 5{,}700, 100, and
2{,}250 --- 0.1--0.3 standard errors---run through a benchmark harness with no
model in the judging role.

The criteria, constants, and reporting plans of every
experiment reported here were committed to version control
before execution --- pre-specified in that sense, not
registered with an external registry. The manuscript passes
an automated set of consistency checks (reported-number, integrity,
and cross-reference checks) and a full-manuscript semantic
audit. The protocol is version 3 of a document whose change-log
records v1$\to$v2 and v2$\to$v3; v1 and v2 are recoverable from
version-control history (Table~\ref{tab:commits}). Table~\ref{tab:commits} gives, for every registered experiment, the commit that
fixed its criteria, its timestamp, and the evidence for the start of its first run.
The code record of the snapshot (version 5, md5 \texttt{f3e17729a294e867968ea1419aa95e72})
includes it, the E-2 disclosure, and every analysis protocol and
result document.

A code-and-configuration snapshot (version 5) accompanies this work as Online Resource 1 and 2, a code record
(1{,}395 files plus \texttt{TREE\_MD5.txt}, md5 \texttt{f3e17729a294e867968ea1419aa95e72}) and a data record (45 files plus \texttt{TREE\_MD5.txt}, md5 \texttt{ddc6e5e58103aa1907fc23ebb4bb07bd}), and is deposited as two Zenodo records
(DOIs \url{https://doi.org/10.5281/zenodo.22710121} and \url{https://doi.org/10.5281/zenodo.22721044}, respectively). The corpora are drawn from public sources (Section~4); the code record includes the collection scripts and the
QA-generation configuration from which the derived sets are regenerated, and the data record provides the generated QA sets, the registered
held-out query sets, the corpus chunk file (with the arXiv-derived text and the verbatim answer sentences omitted and rebuilt by a script in the code record against md5 registries),
the probe and concept sets, and the human label sheet; fifty QA items that ask for personal contact details or author identities are excluded
and listed.

The \kbs{} abbreviation was
introduced late in the project; the released artifacts,
scripts and result files still use the earlier
\texttt{kbs\_*} naming.

\begin{sidewaystable}
\centering
\scriptsize
\caption{Provenance of every registered verdict: evaluation split and size
(queries / documents), correctness criterion, statistic and decision rule, the
registered resampling unit, and the same statistic under document-level cluster
resampling ($B = 10{,}000$, seed 42; the registered query-level values are
reproduced exactly at the same seed). No verdict changes under the cluster unit.
The intervals of Table~\ref{tab:signal_profile} move by at most 0.02 at either
end under document clusters, and the tenfold-budget probabilities of
Table~\ref{tab:b7} become 0.000 and 0.015. Referenced from
Sections~4 and~9.6.}
\label{tab:prov}
\setlength{\tabcolsep}{3pt}
\fitheight{tab:prov}{%
\begin{tabular}{lp{3.4cm}lp{6.6cm}lp{5.0cm}}
\toprule
\textbf{Verdict} & \textbf{Split ($n$ queries / documents)} & \textbf{Criterion} & \textbf{Statistic and rule} & \textbf{Unit} & \textbf{Document-cluster re-analysis} \\
\midrule
R-1 & SciTech test (1,194 / 75); WikiGen test (1,681 / 150) & token-F1 & paired $\Delta$F1, R$\to$F minus Off; $P(>0) > 0.7$ on two of three seeds and R$\to$F above Off on all three & query, $N = 1{,}000$ & pass on Qwen ($P$ 0.87 / 0.98 / 0.92), fail elsewhere; unchanged \\
R-2 & same & token-F1 & paired $\Delta$F1, R$\to$F minus Random-$N$; $P(>0) > 0.7$ on two of three seeds & query, $N = 1{,}000$ & fail on every family (Qwen 0.03 / 0.68 / 0.13); unchanged \\
F-1 & hard-negative test subsets (386 SciTech, 318 WikiGen) & --- & unique preferences $\ge 20$ (count) & none & --- \\
F-2 & same & gold-chunk rank & $\Delta$MRR under preference re-ranking; $P(>0) > 0.7$ & query, $N = 1{,}000$ & not recomputed ($P = 0.000$ as registered) \\
F-3 & same & gold-chunk rank & real minus random-preference $\Delta$MRR $> 0$ & none & --- \\
E-1 & SciTech query set (1,196 / 75); WikiGen query set (1,759 / 150) & containment & $P(\mathrm{acc}_{\text{answered}} > \mathrm{acc}_{\text{abstained}}) > 0.7$, both corpora & query, $N = 10{,}000$ & 1.000 / 1.000; unchanged \\
E-2 & same & containment & risk--coverage AUC of the policy below always-retrieval and below max-confidence, both corpora & none & --- \\
E-3 & same & containment & forced-routing accuracy at $\gamma = 2$ above $\gamma = 0$, $P > 0.95$, both corpora & query, $N = 10{,}000$ & 1.000 / 1.000; unchanged \\
E-4 & same & containment & policy above \kbs{}-only and above max-confidence, $P > 0.95$, both corpora & query, $N = 10{,}000$ & 0.000 / 0.000 against \kbs{}-only; unchanged \\
\bottomrule
\end{tabular}
}
\end{sidewaystable}

\begin{sidewaystable}
\centering
\scriptsize
\caption{Dates and times are given for audit purposes. Version-control provenance of the registered experiments: the commit
that fixed each experiment's criteria (UTC), the evidence for the start of its
first run, and the commit that recorded its result. The routing protocol's
v1$\to$v2 change-log (router formula, abstention-threshold sweep, two
best-case definitions, calibration scope, contrast-corpus exclusion, \kbs{}-only baseline
disclosure, E-3 redefinition) and v2$\to$v3 change-log (raw-rank conflict gate,
threshold replacement, two ablation additions) are recorded in the protocol file
itself; the E-2 evaluation-code correction (per-baseline comparison replacing a
pooled reference) is disclosed in Section~7.}
\label{tab:commits}
\setlength{\tabcolsep}{3pt}
\fitheight{tab:commits}{%
\begin{tabular}{p{5.6cm}p{6.4cm}p{9.6cm}}
\toprule
\textbf{Experiment} & \textbf{Criteria commit (UTC)} & \textbf{First run (evidence); result commit (UTC)} \\
\midrule
Trigger scale-up, LLaMA/SciTech (R, F) & \texttt{aac9c76}, 15 Jul 02:31 & 15 Jul 04:31 (first prediction file; runner committed 05:07, run log ends 06:58); result \texttt{2791ec4}, 15 Jul 07:26 \\
Trigger scale-up, LLaMA/WikiGen & same protocol & 20 Jul 04:32 (first prediction file); result \texttt{664c0d6}, 20 Jul 07:22 \\
Trigger scale-up, Qwen / Mistral / Gemma & same protocol; runner parameterized \texttt{d6d4384}, 22 Jul 00:00 & 22 Jul 09:42 / 22 Jul 14:16 / 23 Jul 00:14 (first prediction files); result \texttt{ea61d3e}, 23 Jul 05:38 \\
Routing pilot (E-1--E-4) & \texttt{da6bb91}, 21 Jul 07:12 (protocol v3) & 21 Jul (CPU, same day); result \texttt{46ebad0}, 21 Jul 08:04 \\
Recalibration invariance (Section~9.4) & \texttt{7769b84}, 21 Aug 14:28 & 21 Aug 14:55; result \texttt{fc985ec}, 21 Aug 15:03 \\
Full fine-tuning control (Section~9.5) & \texttt{625e04a}, 25 Aug 11:07 & 30 Aug 06:52; result \texttt{c4ee2f1}, 30 Aug 09:40 \\
Retrieval-failure and criterion analyses (Section~9.6) & \texttt{029702b}, 31 Aug 03:43; \texttt{ac57cb8}, 2 Sep 00:20; \texttt{10b9d5b}, 2 Sep 08:02 & 31 Aug 04:16; 2 Sep 00:46; 2 Sep (CPU); result \texttt{b5794c2}, 31 Aug 11:38; \texttt{1f0eb15}, 2 Sep 02:19; \texttt{c3f3d11}, 2 Sep 13:24 \\
Signal-profile regeneration (Table~\ref{tab:signal_profile}) & \texttt{f91f41a}, 31 Aug 11:11 & 1 Sep 02:49 (GPU allocation acknowledgement); result \texttt{cf3392a}, 1 Sep 06:10 \\
Tenfold budget (Table~\ref{tab:b7}) & \texttt{34885bd}, 2 Sep 14:13 & 2 Sep 23:26 (GPU allocation acknowledgement); result \texttt{716fd14}, 3 Sep 03:50 \\
Cluster bootstrap (Table~\ref{tab:prov}) & \texttt{babeb3e}, 3 Sep 08:30 & 3 Sep (CPU, same day); result this revision \\
\bottomrule
\end{tabular}
}
\end{sidewaystable}

\begin{table}[htbp]
\centering
\scriptsize
\caption{Expanded transfer budget, one combination (LLaMA/SciTech, seed 42;
exploratory, the registered verdict unchanged). Ten LoRA steps per cycle
on the frozen batch of 32, three cycles, parametric-side signals
regenerated with the current adapter before cycles 2 and 3; the random
control is budget-matched to the realized R$\to$F total (64). $P$ values are
the registered paired query-level bootstrap ($N = 1{,}000$, seed 20260715).}
\label{tab:b7}
\setlength{\tabcolsep}{4pt}
\fitwidth{tab:b7}{%
\begin{tabular}{lccccc}
\toprule
\textbf{Condition} & \textbf{Cycles run} & \textbf{Distilled} & \textbf{Peak $\delta$ per cycle} & \textbf{Rollback} & \textbf{Closed-book F1} \\
\midrule
Undistilled (off, registered) & --- & 0 & --- & --- & 0.1005 \\
R$\to$F, tenfold budget & 2 of 3 & 32 + 32 & 0.093, 0.130 & cycle 2 & 0.078 \\
Random, budget-matched & 3 & 32 + 32 + 0 & 0.074, 0.093, 0.093 & no & 0.086 \\
\midrule
\multicolumn{6}{l}{R$\to$F $-$ off: $-0.023$ [$-0.032$, $-0.013$], $P(>0) = 0.000$; \quad R$\to$F $-$ random: $-0.008$ [$-0.015$, $-0.001$], $P(>0) = 0.009$ ($n = 1{,}194$).} \\
\bottomrule
\end{tabular}%
}
\end{table}

\begin{table}[htbp]
\centering
\scriptsize
\caption{Confidence-term ablation of the R$\to$F selector. Each entry is the
selected batch's target count $k/32$ (retrieval right $\wedge$ parametric wrong)
with the exact hypergeometric $P(X \ge k)$; weights $(w_d, w_\phi, w_p)$ ablate
the three ranked signals of the composite. Random-32 draws and population rates
as in Table~\ref{tab:enrichment}. Discussed in Section~9.2.}
\label{tab:b5abl}
\setlength{\tabcolsep}{3pt}
\fitwidth{tab:b5abl}{%
\begin{tabular}{lccccccc}
\toprule
\textbf{Model / corpus} & \textbf{full} & \textbf{no-$p$} & \textbf{$d{+}p$}
  & \textbf{$\phi{+}p$} & \textbf{$p$-only} & \textbf{$d$-only}
  & \textbf{$\phi$-only} \\
\midrule
LLaMA/Sci & 32 (7.8e-05) & 31 (0.00095) & 22 (0.84) & 31 (0.00095) & 25 (0.42) & 16 (1) & 28 (0.063) \\
Qwen/Sci & 28 (0.021) & 30 (0.0013) & 28 (0.021) & 29 (0.0062) & 22 (0.68) & 21 (0.8) & 28 (0.021) \\
Gemma/Sci & 28 (0.0099) & 27 (0.03) & 23 (0.4) & 27 (0.03) & 24 (0.26) & 24 (0.26) & 32 (3.7e-06) \\
Mistral/Sci & 26 (0.039) & 23 (0.28) & 24 (0.17) & 27 (0.014) & 23 (0.28) & 22 (0.43) & 26 (0.039) \\
LLaMA/Wiki & 28 (0.036) & 26 (0.18) & 29 (0.011) & 30 (0.0027) & 21 (0.86) & 29 (0.011) & 26 (0.18) \\
\bottomrule
\end{tabular}%
}
\vspace{4pt}

\fitwidth{tab:b5abl}{%
\begin{tabular}{lccc}
\toprule
\textbf{F$\to$R} & \textbf{Population target (rate)}
  & \textbf{full $k/32$ ($P \ge k$)} & \textbf{$p$-only $k/32$ ($P \ge k$)} \\
\midrule
LLaMA/Sci & 1 (0.08\%) & 0 (1) & 0 (1) \\
Qwen/Sci & 9 (0.75\%) & 0 (1) & 2 (0.022) \\
Gemma/Sci & 6 (0.50\%) & 0 (1) & 0 (1) \\
Mistral/Sci & 0 (0.00\%) & 0 (1) & 0 (1) \\
LLaMA/Wiki & 5 (0.28\%) & 0 (1) & 0 (1) \\
\bottomrule
\end{tabular}%
}
\end{table}

\begin{table}[htbp]
\centering
\footnotesize
\caption{Base and adapted closed-book F1 per evaluated combination. Base is
the unadapted model on the held-out query set; adapted is the deployed
adapter's record. Two combinations are the base model after guard rollback.}
\label{tab:b6base}
\begin{tabular}{llccc}
\toprule
\textbf{Model} & \textbf{Corpus} & \textbf{Base F1} & \textbf{Adapted F1} & \textbf{$\Delta$F1} \\
\midrule
LLaMA-3.1-8B & SciTech & 0.083 & 0.100 & $+0.017$ \\
Qwen2.5-7B & SciTech & 0.080 & 0.112 & $+0.031$ \\
Gemma-2-9B & SciTech & 0.089 & 0.099 & $+0.010$ \\
Mistral-7B & SciTech & 0.084 & 0.084 & 0 (rollback) \\
LLaMA-3.1-8B & WikiGen & 0.176 & 0.176 & 0 (rollback) \\
\bottomrule
\end{tabular}
\end{table}

\begin{table}[htbp]
\centering
\scriptsize
\caption{Component ablation of Table~\ref{tab:component_ablation}
recomputed under four correctness criteria (four base models, the
2{,}860 gold-span-eligible held-out queries, gates at their
best-case threshold, random gate at the parametric base rate). The
containment rows reproduce the recorded table exactly. FT-unique
counts queries the parametric mode answers and retrieval does not.}
\label{tab:b6crit}
\setlength{\tabcolsep}{3pt}
\fitwidth{tab:b6crit}{%
\begin{tabular}{llcccccccc}
\toprule
\textbf{Model} & \textbf{Criterion} & \textbf{FT-only} & \textbf{RAG-only}
  & \textbf{Best-case} & \textbf{$p_{\mathrm{FT}}$-gate*} & \textbf{$\phi$-gate*}
  & \textbf{Random} & \textbf{RAG/best-case} & \textbf{FT-unique} \\
\midrule
\multirow{4}{*}{LLaMA-3.1-8B}
  & containment & 0.073 & 0.809 & 0.811 & 0.809 & 0.809 & 0.756 & 99.8\% & 5 \\
  & token-F1 $\ge 0.3$ & 0.132 & 0.559 & 0.564 & 0.559 & 0.559 & 0.504 & 99.2\% & 13 \\
  & token-F1 $\ge 0.5$ & 0.099 & 0.544 & 0.548 & 0.544 & 0.544 & 0.502 & 99.2\% & 12 \\
  & exact match & 0.043 & 0.440 & 0.442 & 0.440 & 0.440 & 0.425 & 99.5\% & 6 \\
\midrule
\multirow{4}{*}{Qwen2.5-7B}
  & containment & 0.059 & 0.849 & 0.852 & 0.849 & 0.849 & 0.807 & 99.6\% & 9 \\
  & token-F1 $\ge 0.3$ & 0.119 & 0.752 & 0.767 & 0.752 & 0.752 & 0.679 & 98.0\% & 44 \\
  & token-F1 $\ge 0.5$ & 0.060 & 0.620 & 0.630 & 0.621 & 0.620 & 0.591 & 98.4\% & 29 \\
  & exact match & 0.010 & 0.313 & 0.316 & 0.313 & 0.313 & 0.310 & 99.1\% & 8 \\
\midrule
\multirow{4}{*}{Gemma-2-9B}
  & containment & 0.081 & 0.848 & 0.851 & 0.848 & 0.848 & 0.789 & 99.6\% & 9 \\
  & token-F1 $\ge 0.3$ & 0.146 & 0.650 & 0.681 & 0.650 & 0.650 & 0.577 & 95.5\% & 87 \\
  & token-F1 $\ge 0.5$ & 0.076 & 0.408 & 0.442 & 0.408 & 0.408 & 0.383 & 92.4\% & 96 \\
  & exact match & 0.020 & 0.207 & 0.219 & 0.207 & 0.207 & 0.204 & 94.6\% & 34 \\
\midrule
\multirow{4}{*}{Mistral-7B}
  & containment & 0.058 & 0.735 & 0.740 & 0.735 & 0.735 & 0.699 & 99.3\% & 14 \\
  & token-F1 $\ge 0.3$ & 0.175 & 0.814 & 0.830 & 0.814 & 0.814 & 0.701 & 98.1\% & 46 \\
  & token-F1 $\ge 0.5$ & 0.111 & 0.728 & 0.739 & 0.728 & 0.728 & 0.659 & 98.6\% & 29 \\
  & exact match & 0.034 & 0.484 & 0.488 & 0.484 & 0.484 & 0.472 & 99.2\% & 11 \\
\bottomrule
\end{tabular}%
}
\end{table}

\begin{table}[htbp]
\centering
\scriptsize
\caption{Criterion sensitivity of the finer diagnostics. Panel (a):
target over-representation of the registered full-composite batch (32
queries) when the target is redefined under each criterion --- target
population, selected count, exact hypergeometric $P(X \ge k)$. Panel
(b): $p_{\mathrm{FT}}$ AUROC for parametric correctness on the
retrieval-failure subsets (gold chunk beyond rank 1 or rank 4, ranks
recomputed deterministically), registered criterion and token-F1
$\ge 0.3$, with the number of subset queries the parametric mode
answers and retrieval does not; a dash marks a degenerate subset
without parametric correct answers.}
\label{tab:b6enrich}
\setlength{\tabcolsep}{3pt}
\fitwidth{tab:b6enrich}{%
\begin{tabular}{lcccc}
\toprule
\multicolumn{5}{l}{\emph{(a) Over-representation by criterion: target population / selected $k$ of 32 / $P(X \ge k)$}} \\
\textbf{Model / corpus} & \textbf{containment} & \textbf{token-F1 $\ge 0.3$} & \textbf{token-F1 $\ge 0.5$} & \textbf{exact match} \\
\midrule
LLaMA/SciTech & 894 / 32 / $7.8\times10^{-5}$ & 42 / 2 / 0.31 & 38 / 1 / 0.65 & 0 / 0 / --- \\
Qwen/SciTech & 845 / 28 / 0.021 & 737 / 24 / 0.079 & 620 / 17 / 0.51 & 356 / 10 / 0.49 \\
Gemma/SciTech & 814 / 28 / 0.0099 & 692 / 15 / 0.93 & 602 / 13 / 0.90 & 364 / 4 / 1.00 \\
Mistral/SciTech & 783 / 26 / 0.039 & 732 / 18 / 0.78 & 656 / 13 / 0.97 & 362 / 7 / 0.90 \\
LLaMA/WikiGen & 1{,}275 / 28 / 0.036 & 1{,}249 / 28 / 0.024 & 1{,}299 / 28 / 0.051 & 1{,}140 / 26 / 0.033 \\
\midrule
\multicolumn{5}{l}{\emph{(b) Retrieval-failure subsets: AUROC (FT-right-RAG-wrong count)}} \\
\textbf{Model / corpus} & \textbf{rank $>1$, containment} & \textbf{rank $>1$, token-F1} & \textbf{rank $>4$, containment} & \textbf{rank $>4$, token-F1} \\
\midrule
LLaMA/SciTech ($n$ = 614 / 183) & 0.676 (1) & 0.628 (1) & 0.623 (1) & 0.678 (1) \\
Qwen/SciTech (614 / 183) & 0.785 (6) & 0.551 (24) & 0.852 (5) & 0.515 (18) \\
Gemma/SciTech (614 / 183) & 0.673 (3) & 0.558 (23) & 0.753 (1) & 0.548 (12) \\
Mistral/SciTech (614 / 183) & 0.863 (0) & 0.550 (19) & 0.807 (0) & 0.476 (11) \\
LLaMA/WikiGen (490 / 69) & 0.803 (2) & 0.725 (6) & --- (0) & 0.537 (2) \\
\bottomrule
\end{tabular}%
}
\end{table}

\begin{table}[htbp]
\centering
\footnotesize
\caption{Routing and abstention pilot results. Sel.\ acc.\ is
selective accuracy over answered queries, Cov.\ is coverage,
RC-AUC is the area under the risk--coverage curve (lower is
better), ECE is expected calibration error. Bold marks the best
ECE per corpus. Best-case rows are upper bounds chosen with knowledge of the answers, not
deployable policies. Plotted as Figure~\ref{fig:ensemble};
discussed in Section~7.}
\label{tab:ensemble_results}
\setlength{\tabcolsep}{5pt}
\begin{tabular}{lcccc}
\toprule
\textbf{Policy} & \textbf{Sel.} & \textbf{Cov.}
  & \textbf{RC-} & \textbf{ECE} \\
 & \textbf{acc.} & & \textbf{AUC} & \\
\midrule
\multicolumn{5}{l}{\emph{(a) SciTech} ($n=1{,}196$, \kbs{} $=0.164$)}\\
Always-RAG        & 0.806 & 1.000 & 0.078 & 0.306 \\
Always-FT         & 0.059 & 1.000 & 0.902 & 0.441 \\
Max-confidence    & 0.467 & 1.000 & 0.483 & 0.243 \\
\kbs{}-only       & 0.806 & 1.000 & 0.078 & 0.306 \\
Ensemble          & 0.725 & 0.809 & 0.159 & \textbf{0.136} \\
\midrule
Best-case routing & 0.807 & 1.000 & 0.020 & 0.000 \\
Best-case abstention & 1.000 & 0.807 & 0.020 & 0.000 \\
\midrule
\multicolumn{5}{l}{\emph{(b) WikiGen} ($n=1{,}759$, \kbs{} $=0.244$)}\\
Always-RAG        & 0.814 & 1.000 & 0.155 & 0.332 \\
Always-FT         & 0.092 & 1.000 & 0.789 & 0.408 \\
Max-confidence    & 0.485 & 1.000 & 0.431 & 0.232 \\
\kbs{}-only       & 0.814 & 1.000 & 0.155 & 0.332 \\
Ensemble          & 0.772 & 0.756 & 0.178 & \textbf{0.245} \\
\midrule
Best-case routing & 0.817 & 1.000 & 0.018 & 0.000 \\
Best-case abstention & 1.000 & 0.817 & 0.018 & 0.000 \\
\bottomrule
\end{tabular}
\end{table}

\begin{figure}[htbp]
  \centering
  \includegraphics[width=\linewidth]{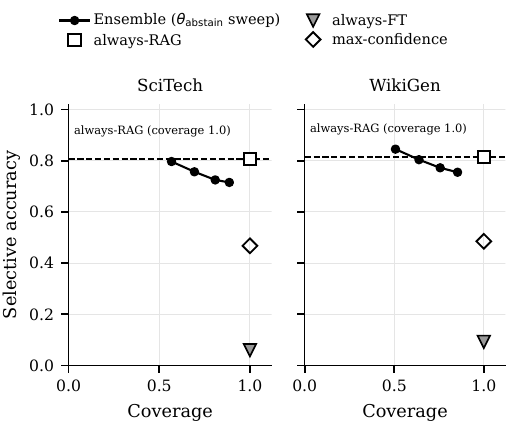}
  \caption{Coverage and selective accuracy of the routing
policy. Baselines: always-RAG answers every query with
retrieval; max-confidence routes each query to the mode
with the higher raw score; always-FT answers
parametrically.}
  \label{fig:ensemble}
\end{figure}

\FloatBarrier
\section{Corrected-Template Replication of the LLaMA/SciTech Combination}
\label{sup:b9}

The LLaMA/SciTech held-out query set was regenerated under the corrected prompt
template (Section~4) and the trigger experiment was re-run on it
with the registered protocol --- three seeds, R$\to$F against Off and against
Random-$N$, token-F1 on the held-out test set --- as a post-registration
replication; the routing pilot was re-run with the regenerated SciTech held-out
query set and the registered WikiGen set. The registered artifact and its verdicts
stand as recorded; the replication is reported beside them and does not replace
them. Table~\ref{tab:b9} gives the trigger rows with the registered
bootstrap form ($N = 1{,}000$) and the query- and document-cluster resampling of
Table~\ref{tab:prov}; Table~\ref{tab:b9_pilot} the routing pilot; and
Table~\ref{tab:b9_diag} the closed-book diagnostics of the two held-out query sets.
Under the corrected prompt template R-1 fails and R-2 passes ($P > 0.7$ on two of three
seeds against Random-$N$); the registered record fails both.

\begin{table}[htbp]
\centering
\scriptsize
\caption{Trigger replication, LLaMA/SciTech: recorded (pre-fix) held-out query set
recomputed beside the regenerated one. F1 is closed-book token-F1 on the test set;
$P$ is the one-sided bootstrap probability that R$\to$F exceeds the comparator,
in the registered form ($N = 1{,}000$) / by query / by document cluster
($B = 10{,}000$). In the verdict rows, `$P > 0.7$ on $x$ / $y$' counts the seeds
clearing the threshold against Off / against Random-$N$.}
\label{tab:b9}
\setlength{\tabcolsep}{4pt}
\fitwidth{tab:b9}{%
\begin{tabular}{llccccc}
\toprule
\textbf{Query set} & \textbf{Seed} & \textbf{F1 Off} & \textbf{F1 R$\to$F} & \textbf{F1 Random-$N$} & \textbf{$P$(R$\to$F $>$ Off)} & \textbf{$P$(R$\to$F $>$ Random-$N$)} \\
\midrule
recorded & 42 & 0.1005 & 0.0959 & 0.1023 & 0.141 / 0.149 / 0.157 & 0.007 / 0.007 / 0.007 \\
recorded & 43 & 0.1005 & 0.0961 & 0.1040 & 0.162 / 0.160 / 0.182 & 0.003 / 0.003 / 0.007 \\
recorded & 44 & 0.1005 & 0.0959 & 0.1024 & 0.142 / 0.149 / 0.172 & 0.025 / 0.033 / 0.045 \\
\addlinespace
regenerated & 42 & 0.1005 & 0.1017 & 0.0971 & 0.656 / 0.640 / 0.637 & 0.918 / 0.914 / 0.903 \\
regenerated & 43 & 0.1005 & 0.1019 & 0.0989 & 0.692 / 0.666 / 0.656 & 0.840 / 0.821 / 0.822 \\
regenerated & 44 & 0.1005 & 0.1010 & 0.1005 & 0.593 / 0.568 / 0.564 & 0.606 / 0.573 / 0.567 \\
\midrule
\multicolumn{2}{l}{verdict, recorded} & \multicolumn{5}{l}{R-1 fail, R-2 fail (R$\to$F above Off on 0 seeds, above Random-$N$ on 0; $P > 0.7$ on 0 / 0)} \\
\multicolumn{2}{l}{verdict, regenerated} & \multicolumn{5}{l}{R-1 fail, R-2 pass (R$\to$F above Off on 3 seeds, above Random-$N$ on 3; $P > 0.7$ on 0 / 2)} \\
\bottomrule
\end{tabular}%
}
\end{table}

\begin{sidewaystable}
\centering
\scriptsize
\caption{Routing pilot on the recorded and the regenerated SciTech held-out query set
(WikiGen registered in both). Containment criterion; E-3 $P$ is given for
SciTech / WikiGen.}
\label{tab:b9_pilot}
\setlength{\tabcolsep}{4pt}
\fitheight{tab:b9_pilot}{%
\begin{tabular}{lccccccc}
\toprule
\textbf{Query set} & \textbf{E-1 / E-2 / E-3 / E-4} & \textbf{E-3 $P$} & \textbf{RAG acc.} & \textbf{FT acc.} & \textbf{Policy sel.\ acc.\ (cov.)} & \textbf{RC-AUC policy / always-RAG} & \textbf{Decisions rag / ft / abstain} \\
\midrule
recorded & pass / fail / pass / fail & 1.000 / 1.000 & 0.806 & 0.059 & 0.725 (0.81) & 0.159 / 0.078 & 754 / 213 / 229 \\
regenerated & pass / fail / pass / fail & 1.000 / 1.000 & 0.750 & 0.051 & 0.648 (0.80) & 0.238 / 0.164 & 744 / 212 / 240 \\
\bottomrule
\end{tabular}
}
\end{sidewaystable}

\begin{sidewaystable}
\centering
\scriptsize
\caption{Closed-book diagnostics of the LLaMA/SciTech held-out query set, recorded
(pre-fix) and regenerated ($n = 1{,}196$). Correctness is containment unless
stated; the confidence signal is $p_{\mathrm{FT}}$.}
\label{tab:b9_diag}
\setlength{\tabcolsep}{4pt}
\fitheight{tab:b9_diag}{%
\begin{tabular}{lccccccccc}
\toprule
\textbf{Query set} & \textbf{AUROC} & \textbf{corr} & \textbf{ECE} & \textbf{FT acc.} & \textbf{FT acc.\ (F1 $\ge 0.3$)} & \textbf{FT F1 mean} & \textbf{RAG acc.} & \textbf{Answer words (median)} & \textbf{Unterminated answers} \\
\midrule
recorded & 0.649 & $+0.120$ & 0.686 & 0.059 & 0.016 & 0.038 & 0.806 & 45 & 1{,}193 \\
regenerated & 0.602 & $+0.088$ & 0.487 & 0.051 & 0.131 & 0.104 & 0.750 & 7 & 0 \\
\bottomrule
\end{tabular}
}
\end{sidewaystable}

\FloatBarrier
\section{Routing on Retrieval-Failure Subsets}
\label{sup:b11}

The routing pilot of Section~7 re-evaluated on the held-out
queries whose gold chunk is retrieved beyond rank 1 and beyond rank 4 (ranks from
the deterministic bge-m3 pass of Section~9.6). The deployed
policy, its thresholds, the corpus-level \kbs{}, and the percentile ranks are those
of the full held-out query set; only the evaluation is restricted. The hard-negative
subsets of the trigger experiment are test-split queries and do not intersect the
held-out query sets. The full-set rows reproduce the registered pilot; the regenerated
SciTech rows use the corrected-template held-out query set (Section~\ref{sup:b9}).
Bootstrap $P$ values use $B = 10{,}000$ at the registered seed.

\begin{sidewaystable}
\centering
\scriptsize
\caption{Routing pilot on retrieval-failure subsets. Selective accuracy /
risk--coverage area per baseline (ensemble with its coverage); possible gain =
best-case-routing accuracy minus always-retrieval; FR$_{\gamma=2}$ = forced-routing
accuracy at coverage 1; E-2 by the intent interpretation (ensemble RC-AUC below
always-retrieval and below max-confidence); E-4 as $P(\mathrm{FR}_{\gamma=2} >
\kbs\text{-only})$ / $P(\mathrm{FR}_{\gamma=2} > \text{max-confidence})$.}
\label{tab:b11}
\setlength{\tabcolsep}{3pt}
\fitheight{tab:b11}{%
\begin{tabular}{lllccccccccccc}
\toprule
\textbf{Block} & \textbf{Corpus} & \textbf{Subset} & $n$ & \textbf{RAG acc.} & \textbf{FT acc.} & \textbf{Always-RAG} & \textbf{Max-conf.} & \textbf{Ensemble (cov.)} & \textbf{Best-case} & \textbf{Possible gain} & \textbf{FR$_{\gamma=2}$} & \textbf{E-2} & \textbf{E-4 $P$} \\
\midrule
registered & SciTech & all & 1196 & 0.806 & 0.059 & 0.806 / 0.078 & 0.467 / 0.483 & 0.725 (0.81) / 0.159 & 0.807 & +0.001 & 0.684 & fail & 0.000 / 1.000 \\
registered & SciTech & rank $>1$ & 614 & 0.707 & 0.050 & 0.707 / 0.120 & 0.401 / 0.533 & 0.684 (0.76) / 0.211 & 0.708 & +0.002 & 0.609 & fail & 0.000 / 1.000 \\
registered & SciTech & rank $>4$ & 183 & 0.224 & 0.055 & 0.224 / 0.586 & 0.115 / 0.844 & 0.268 (0.53) / 0.679 & 0.230 & +0.005 & 0.191 & fail & 0.204 / 0.974 \\
\addlinespace
registered & WikiGen & all & 1759 & 0.814 & 0.092 & 0.814 / 0.155 & 0.485 / 0.431 & 0.772 (0.76) / 0.178 & 0.817 & +0.003 & 0.746 & fail & 0.000 / 1.000 \\
registered & WikiGen & rank $>1$ & 490 & 0.745 & 0.084 & 0.745 / 0.219 & 0.453 / 0.449 & 0.712 (0.77) / 0.242 & 0.749 & +0.004 & 0.667 & fail & 0.003 / 1.000 \\
registered & WikiGen & rank $>4$ & 69 & 0.217 & 0.000 & 0.217 / 0.769 & 0.130 / 0.879 & 0.184 (0.71) / 0.795 & 0.217 & +0.000 & 0.174 & fail & 0.227 / 0.727 \\
\addlinespace
regenerated SciTech & SciTech & all & 1196 & 0.750 & 0.051 & 0.750 / 0.164 & 0.431 / 0.524 & 0.647 (0.80) / 0.238 & 0.756 & +0.006 & 0.620 & fail & 0.000 / 1.000 \\
regenerated SciTech & SciTech & rank $>1$ & 614 & 0.656 & 0.041 & 0.656 / 0.227 & 0.360 / 0.591 & 0.559 (0.78) / 0.316 & 0.661 & +0.005 & 0.531 & fail & 0.000 / 1.000 \\
regenerated SciTech & SciTech & rank $>4$ & 183 & 0.230 & 0.049 & 0.230 / 0.689 & 0.131 / 0.829 & 0.195 (0.67) / 0.754 & 0.240 & +0.011 & 0.180 & fail & 0.106 / 0.890 \\
\bottomrule
\end{tabular}
}
\end{sidewaystable}

\FloatBarrier
\section{Selector Ablation of the Trigger's Downstream Outcome}
\label{sup:b10}

The registered transfer (three cycles of one LoRA step on a batch of 32, learning
rate $10^{-4}$, rate cap as registered) re-run on the corrected-template
LLaMA/SciTech held-out query set from batches chosen by the full composite and by
four ablated selectors, each on seeds 42--44; the full-selector, random-batch, and
undistilled rows are the corrected-template replication of
Section~\ref{sup:b9}. $P$ is the one-sided paired bootstrap probability that
the selected batch exceeds the comparator on closed-book test F1, in the registered
form ($N = 1{,}000$) / by query / by document cluster ($B = 10{,}000$). The
selected batch is deterministic (identical top-32 in every cycle); its target
count is the number of the 32 that are retrieval-right and parametric-wrong
under containment (population rate 0.705; random-batch median 23, 2.5--97.5th
percentiles 17--27). By the protocol's pre-fixed rule the full-versus-no-confidence
comparison is \emph{mixed}: the paired difference lies inside the bootstrap width
on two seeds and outside it on one, where the selector without the confidence
term does better.

\begin{sidewaystable}
\centering
\scriptsize
\caption{Selector ablation, LLaMA/SciTech, corrected prompt template. F1 = undistilled /
selected batch / random batch; $P$ = registered form / query / document cluster;
the full $-$ no-p column is the paired contrast behind `no statistically detectable
benefit'.}
\label{tab:b10}
\setlength{\tabcolsep}{3pt}
\fitheight{tab:b10}{%
\begin{tabular}{llcccccc}
\toprule
\textbf{Selector} & \textbf{Weights $(d, \phi, 1-p)$} & \textbf{Seed} & \textbf{F1 off / selected / random} & \textbf{$P$(selected $>$ off)} & \textbf{$P$(selected $>$ random)} & \textbf{Target in batch} & \textbf{full $-$ no-p $\Delta$F1 [CI query] / [CI cluster]} \\
\midrule
full (registered selector) & (1, 1, 1) & 42 & 0.1005 / 0.1017 / 0.0971 & 0.656 / 0.640 / 0.637 & 0.918 / 0.914 / 0.903 & 26/32 & --- \\
full (registered selector) & (1, 1, 1) & 43 & 0.1005 / 0.1019 / 0.0989 & 0.692 / 0.666 / 0.656 & 0.840 / 0.821 / 0.822 & 26/32 & --- \\
full (registered selector) & (1, 1, 1) & 44 & 0.1005 / 0.1010 / 0.1005 & 0.593 / 0.568 / 0.564 & 0.606 / 0.573 / 0.567 & 26/32 & --- \\
\addlinespace
without confidence & (1, 1, 0) & 42 & 0.1005 / 0.1058 / 0.0971 & 0.959 / 0.945 / 0.927 & 0.995 / 0.994 / 0.987 & 25/32 & $-0.004$ [$-0.009$, $+0.001$] / [$-0.009$, $+0.001$] \\
without confidence & (1, 1, 0) & 43 & 0.1005 / 0.1075 / 0.0989 & 0.992 / 0.980 / 0.974 & 0.998 / 0.995 / 0.993 & 25/32 & $-0.006$ [$-0.011$, $-0.001$] / [$-0.011$, $-0.001$] \\
without confidence & (1, 1, 0) & 44 & 0.1005 / 0.1049 / 0.1005 & 0.931 / 0.907 / 0.893 & 0.940 / 0.938 / 0.951 & 25/32 & $-0.004$ [$-0.009$, $+0.001$] / [$-0.009$, $+0.001$] \\
\addlinespace
confidence alone & (0, 0, 1) & 42 & 0.1005 / 0.0999 / 0.0971 & 0.441 / 0.438 / 0.444 & 0.830 / 0.834 / 0.849 & 26/32 & --- \\
confidence alone & (0, 0, 1) & 43 & 0.1005 / 0.0993 / 0.0989 & 0.382 / 0.374 / 0.380 & 0.569 / 0.536 / 0.551 & 26/32 & --- \\
confidence alone & (0, 0, 1) & 44 & 0.1005 / 0.0995 / 0.1005 & 0.416 / 0.394 / 0.398 & 0.375 / 0.379 / 0.379 & 26/32 & --- \\
\addlinespace
context similarity alone & (0, 1, 0) & 42 & 0.1005 / 0.0988 / 0.0971 & 0.351 / 0.351 / 0.355 & 0.666 / 0.661 / 0.646 & 26/32 & --- \\
context similarity alone & (0, 1, 0) & 43 & 0.1005 / 0.0990 / 0.0989 & 0.361 / 0.360 / 0.367 & 0.495 / 0.507 / 0.507 & 26/32 & --- \\
context similarity alone & (0, 1, 0) & 44 & 0.1005 / 0.0997 / 0.1005 & 0.423 / 0.430 / 0.431 & 0.408 / 0.421 / 0.427 & 26/32 & --- \\
\addlinespace
disagreement alone & (1, 0, 0) & 42 & 0.1005 / 0.0988 / 0.0971 & 0.263 / 0.276 / 0.299 & 0.684 / 0.701 / 0.686 & 20/32 & --- \\
disagreement alone & (1, 0, 0) & 43 & 0.1005 / 0.0977 / 0.0989 & 0.151 / 0.165 / 0.202 & 0.354 / 0.346 / 0.368 & 20/32 & --- \\
disagreement alone & (1, 0, 0) & 44 & 0.1005 / 0.0990 / 0.1005 & 0.306 / 0.312 / 0.340 & 0.284 / 0.285 / 0.282 & 20/32 & --- \\
\bottomrule
\end{tabular}%
}
\end{sidewaystable}

\FloatBarrier
\section{Selector Ablation on the Qwen and Gemma Combinations}
\label{sup:b12}

The ablation of Section~\ref{sup:b10} repeated on the two other genuinely
adapted combinations, Qwen-2.5-7B and Gemma-2-9B on SciTech, from their registered
held-out query sets (generated under the corrected prompt template) and registered starting
adapters; the full-selector, random-batch, and undistilled rows are the
registered scale-up rows of Table~1. Columns, bootstrap
forms, and the target count are as in Table~\ref{tab:b10}
(Table~\ref{tab:b12}). Table~\ref{tab:b12_rules} applies the registered R-1 and
R-2 rules to each ablated selector; the registered composite passes R-1 only on
Qwen and neither on Gemma.

\begin{sidewaystable}
\centering
\scriptsize
\caption{Selector ablation, Qwen-2.5-7B and Gemma-2-9B on SciTech. F1 = undistilled /
selected batch / random batch; $P$ = registered form / query / document cluster;
target = retrieval-right and parametric-wrong queries among the selected 32; the
full $-$ no-p column is the paired contrast behind `no statistically detectable
benefit'.}
\label{tab:b12}
\setlength{\tabcolsep}{3pt}
\fitheight{tab:b12}{%
\begin{tabular}{llcccccc}
\toprule
\textbf{Selector} & \textbf{Weights $(d, \phi, 1-p)$} & \textbf{Seed} & \textbf{F1 off / selected / random} & \textbf{$P$(selected $>$ off)} & \textbf{$P$(selected $>$ random)} & \textbf{Target in batch} & \textbf{full $-$ no-p $\Delta$F1 [CI query] / [CI cluster]} \\
\midrule
\multicolumn{8}{l}{\textit{Qwen-2.5-7B / SciTech (population target rate 0.707; random-batch median 23, 2.5--97.5th percentiles 18--27)}} \\
full (registered selector) & (1, 1, 1) & 42 & 0.1116 / 0.1142 / 0.1189 & 0.848 / 0.844 / 0.870 & 0.048 / 0.040 / 0.026 & 28/32 & --- \\
full (registered selector) & (1, 1, 1) & 43 & 0.1116 / 0.1159 / 0.1145 & 0.946 / 0.949 / 0.977 & 0.677 / 0.660 / 0.677 & 28/32 & --- \\
full (registered selector) & (1, 1, 1) & 44 & 0.1116 / 0.1145 / 0.1178 & 0.874 / 0.866 / 0.915 & 0.149 / 0.140 / 0.129 & 28/32 & --- \\
\addlinespace
without confidence & (1, 1, 0) & 42 & 0.1116 / 0.1165 / 0.1189 & 0.947 / 0.941 / 0.944 & 0.210 / 0.203 / 0.201 & 30/32 & $-0.002$ [$-0.007$, $+0.003$] / [$-0.007$, $+0.003$] \\
without confidence & (1, 1, 0) & 43 & 0.1116 / 0.1152 / 0.1145 & 0.872 / 0.875 / 0.886 & 0.630 / 0.598 / 0.596 & 30/32 & $+0.001$ [$-0.005$, $+0.006$] / [$-0.004$, $+0.005$] \\
without confidence & (1, 1, 0) & 44 & 0.1116 / 0.1157 / 0.1178 & 0.921 / 0.920 / 0.917 & 0.223 / 0.206 / 0.181 & 30/32 & $-0.001$ [$-0.006$, $+0.004$] / [$-0.006$, $+0.004$] \\
\addlinespace
confidence alone & (0, 0, 1) & 42 & 0.1116 / 0.1146 / 0.1189 & 0.869 / 0.861 / 0.857 & 0.088 / 0.078 / 0.084 & 22/32 & --- \\
confidence alone & (0, 0, 1) & 43 & 0.1116 / 0.1129 / 0.1145 & 0.692 / 0.679 / 0.672 & 0.374 / 0.325 / 0.345 & 22/32 & --- \\
confidence alone & (0, 0, 1) & 44 & 0.1116 / 0.1131 / 0.1178 & 0.726 / 0.715 / 0.714 & 0.098 / 0.083 / 0.099 & 22/32 & --- \\
\addlinespace
context similarity alone & (0, 1, 0) & 42 & 0.1116 / 0.1154 / 0.1189 & 0.918 / 0.925 / 0.945 & 0.129 / 0.116 / 0.106 & 28/32 & --- \\
context similarity alone & (0, 1, 0) & 43 & 0.1116 / 0.1172 / 0.1145 & 0.984 / 0.983 / 0.990 & 0.798 / 0.779 / 0.765 & 28/32 & --- \\
context similarity alone & (0, 1, 0) & 44 & 0.1116 / 0.1163 / 0.1178 & 0.971 / 0.965 / 0.983 & 0.343 / 0.310 / 0.312 & 28/32 & --- \\
\addlinespace
disagreement alone & (1, 0, 0) & 42 & 0.1116 / 0.1152 / 0.1189 & 0.855 / 0.859 / 0.886 & 0.136 / 0.129 / 0.104 & 21/32 & --- \\
disagreement alone & (1, 0, 0) & 43 & 0.1116 / 0.1172 / 0.1145 & 0.956 / 0.965 / 0.982 & 0.858 / 0.836 / 0.832 & 21/32 & --- \\
disagreement alone & (1, 0, 0) & 44 & 0.1116 / 0.1204 / 0.1178 & 0.997 / 0.998 / 0.998 & 0.822 / 0.835 / 0.840 & 21/32 & --- \\
\addlinespace
\multicolumn{8}{l}{\textit{Gemma-2-9B / SciTech (population target rate 0.681; random-batch median 22, 2.5--97.5th percentiles 17--27)}} \\
full (registered selector) & (1, 1, 1) & 42 & 0.0990 / 0.0965 / 0.0987 & 0.259 / 0.237 / 0.273 & 0.267 / 0.288 / 0.301 & 28/32 & --- \\
full (registered selector) & (1, 1, 1) & 43 & 0.0990 / 0.0974 / 0.0964 & 0.312 / 0.320 / 0.332 & 0.629 / 0.611 / 0.608 & 28/32 & --- \\
full (registered selector) & (1, 1, 1) & 44 & 0.0990 / 0.0977 / 0.0967 & 0.337 / 0.342 / 0.364 & 0.612 / 0.592 / 0.585 & 28/32 & --- \\
\addlinespace
without confidence & (1, 1, 0) & 42 & 0.0990 / 0.1026 / 0.0987 & 0.884 / 0.884 / 0.864 & 0.881 / 0.893 / 0.886 & 27/32 & $-0.006$ [$-0.012$, $-0.000$] / [$-0.012$, $+0.000$] \\
without confidence & (1, 1, 0) & 43 & 0.0990 / 0.1019 / 0.0964 & 0.849 / 0.830 / 0.823 & 0.941 / 0.948 / 0.941 & 27/32 & $-0.004$ [$-0.010$, $+0.001$] / [$-0.010$, $+0.001$] \\
without confidence & (1, 1, 0) & 44 & 0.0990 / 0.1031 / 0.0967 & 0.910 / 0.907 / 0.879 & 0.967 / 0.960 / 0.955 & 27/32 & $-0.005$ [$-0.011$, $+0.000$] / [$-0.012$, $+0.001$] \\
\addlinespace
confidence alone & (0, 0, 1) & 42 & 0.0990 / 0.0910 / 0.0987 & 0.013 / 0.012 / 0.021 & 0.022 / 0.021 / 0.023 & 24/32 & --- \\
confidence alone & (0, 0, 1) & 43 & 0.0990 / 0.0910 / 0.0964 & 0.013 / 0.013 / 0.023 & 0.063 / 0.077 / 0.063 & 24/32 & --- \\
confidence alone & (0, 0, 1) & 44 & 0.0990 / 0.0888 / 0.0967 & 0.001 / 0.002 / 0.005 & 0.005 / 0.011 / 0.013 & 24/32 & --- \\
\addlinespace
context similarity alone & (0, 1, 0) & 42 & 0.0990 / 0.0996 / 0.0987 & 0.572 / 0.579 / 0.574 & 0.611 / 0.617 / 0.610 & 32/32 & --- \\
context similarity alone & (0, 1, 0) & 43 & 0.0990 / 0.1019 / 0.0964 & 0.798 / 0.808 / 0.782 & 0.910 / 0.909 / 0.922 & 32/32 & --- \\
context similarity alone & (0, 1, 0) & 44 & 0.0990 / 0.1011 / 0.0967 & 0.748 / 0.746 / 0.724 & 0.885 / 0.890 / 0.864 & 32/32 & --- \\
\addlinespace
disagreement alone & (1, 0, 0) & 42 & 0.0990 / 0.0985 / 0.0987 & 0.470 / 0.444 / 0.454 & 0.506 / 0.481 / 0.486 & 24/32 & --- \\
disagreement alone & (1, 0, 0) & 43 & 0.0990 / 0.0991 / 0.0964 & 0.515 / 0.508 / 0.504 & 0.804 / 0.805 / 0.790 & 24/32 & --- \\
disagreement alone & (1, 0, 0) & 44 & 0.0990 / 0.0993 / 0.0967 & 0.555 / 0.541 / 0.530 & 0.786 / 0.777 / 0.736 & 24/32 & --- \\
\bottomrule
\end{tabular}%
}
\end{sidewaystable}

\begin{table}[htbp]
\centering
\scriptsize
\caption{The registered R-1 and R-2 rules applied to each ablated selector
(R-1: selected $>$ undistilled on three seeds and $P > 0.7$ on at least two;
R-2: selected $>$ random on at least two seeds and $P > 0.7$ on at least two;
registered form). The registered composite: Qwen R-1 pass, R-2 fail; Gemma
R-1 fail, R-2 fail.}
\label{tab:b12_rules}
\setlength{\tabcolsep}{5pt}
\fitwidth{tab:b12_rules}{%
\begin{tabular}{llcccc}
\toprule
\textbf{Family} & \textbf{Selector} & \textbf{R-1} & \textbf{R-2} & \textbf{Seeds selected $>$ off / $>$ random} & \textbf{Seeds $P > 0.7$ vs off / vs random} \\
\midrule
Qwen-2.5-7B & without confidence & pass & fail & 3 / 1 & 3 / 0 \\
Qwen-2.5-7B & confidence alone & pass & fail & 3 / 0 & 2 / 0 \\
Qwen-2.5-7B & context similarity alone & pass & fail & 3 / 1 & 3 / 1 \\
Qwen-2.5-7B & disagreement alone & pass & pass & 3 / 2 & 3 / 2 \\
Gemma-2-9B & without confidence & pass & pass & 3 / 3 & 3 / 3 \\
Gemma-2-9B & confidence alone & fail & fail & 0 / 0 & 0 / 0 \\
Gemma-2-9B & context similarity alone & pass & pass & 3 / 3 & 2 / 2 \\
Gemma-2-9B & disagreement alone & fail & pass & 2 / 2 & 0 / 2 \\
\bottomrule
\end{tabular}
}
\end{table}

\FloatBarrier

\end{document}